\documentclass[10pt,twocolumn,letterpaper]{article}

\usepackage[accsupp]{axessibility}  
\usepackage[pagenumbers]{iccv} 

\definecolor{iccvblue}{rgb}{0.21,0.49,0.74}
\usepackage[pagebackref,breaklinks,colorlinks,allcolors=iccvblue]{hyperref}

\usepackage[ruled,noline,nofillcomment]{algorithm2e}
\SetAlCapHSkip{0pt} 
\usepackage{arydshln}
\usepackage{indentfirst} 
\usepackage{tcolorbox}
\usepackage{multirow}
\definecolor{lightgray}{gray}{0.9} 

\definecolor{highLevelColor}{HTML}{EC8877} 
\definecolor{lowLevelColor}{HTML}{6A99D0}  

\definecolor{t2ienhance}{rgb}{0.165, 0.616, 0.561}

\definecolor{noisequerycolor}{rgb}{0.7, 0.5, 0.2} 

\newcommand{\wry}[1]{\textcolor{black}{#1}}
\newcommand{\hhy}[1]{\textcolor{black}{#1}}

\def\paperID{6696} 
\def\confName{ICCV}
\def\confYear{2025}

\title{The Silent Assistant: \emph{NoiseQuery} as Implicit Guidance for \\Goal-Driven Image Generation}

\author{
Ruoyu Wang\textsuperscript{1} \and
Huayang Huang\textsuperscript{1} \and
Ye Zhu\textsuperscript{2} \and 
Olga Russakovsky\textsuperscript{2} \and
Yu Wu\textsuperscript{1}\thanks{Corresponding author.} \and
\\
\textsuperscript{1}School of Computer Science, Wuhan University \\
\textsuperscript{2}Department of Computer Science, Princeton University \\
{\tt\small \{wangruoyu, hyhuang, wuyucs\}@whu.edu.cn} \\
{\tt\small \{yezhu, olgarus\}@princeton.edu}
}

\begin{document}
\maketitle
\addtocontents{toc}{\protect\setcounter{tocdepth}{0}}

\begin{abstract}


In this work, we introduce \textbf{NoiseQuery} as a novel method for enhanced noise initialization in versatile goal-driven text-to-image (T2I) generation. Specifically, we propose to leverage an aligned Gaussian noise as implicit guidance to complement explicit user-defined inputs, such as text prompts, for better generation quality and controllability. Unlike existing noise optimization methods designed for specific models, our approach is grounded in a fundamental examination of the generic finite-step noise scheduler design in diffusion formulation, allowing better generalization across different diffusion-based architectures in a \textbf{tuning-free manner}. This model-agnostic nature allows us to construct a reusable noise library compatible with multiple T2I models and enhancement techniques, serving as a foundational layer for more effective generation.
Extensive experiments demonstrate that \textbf{NoiseQuery} enables fine-grained control and yields significant performance boosts not only over high-level semantics but also over \textbf{low-level visual attributes}, which are typically difficult to specify through text alone, with seamless integration into current workflows with minimal computational overhead. Code is available at \href{https://github.com/wangruoyu02/NoiseQuery}{\small\texttt{https://github.com/wangruoyu02/NoiseQuery}}.

\end{abstract}    
\section{Introduction}
\label{sec:intro}

\begin{figure}[t]
    \centering
    \includegraphics[width=\linewidth]{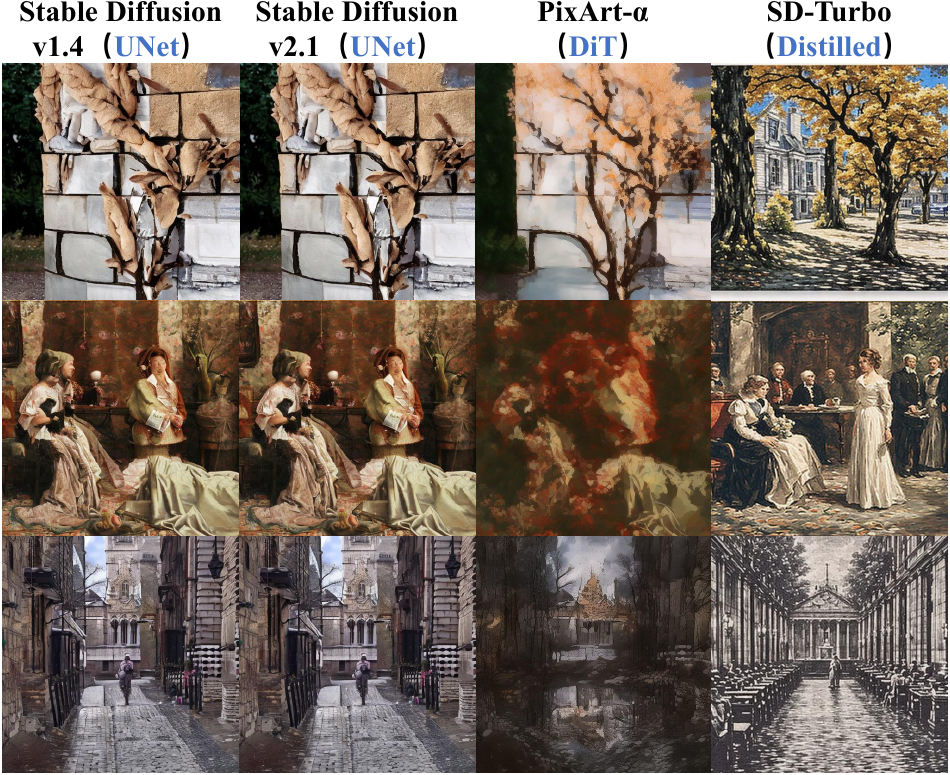}
    \vspace{-5mm}
    \caption{
    Each row shows a set of images generated by different diffusion models starting from the same initial noise, using the same sampler, and with NULL as the text prompt. 
    The generated images demonstrate striking \emph{both high-level semantic and low-level visual similarities} across models.}
    \label{fig:teaser_fig-a}
    \vspace{-4mm}
    \end{figure}


Text-to-image synthesis (T2I) has gained significant research attention with the rising popularity of diffusion generative models~\cite{ho2020denoising-ddpm,song2020denoising-ddim,song2020score,peebles2023scalable-DiT}. 
While these models demonstrate impressive generative capabilities, they still struggle to faithfully align generated content with user prompts, leading to suboptimal outputs. 
To enrich the text-based conditioning capabilities of diffusion models, researchers have explored various techniques, including fine-tuning diffusion UNet~\cite{fan2023dpok,wallace2024diffusion-dpo,black2023training-rlhf}, enhancing language encoders~\cite{zhao2024bridging-lavi,hu2024ella}, refining inference strategies~\cite{chefer2023attend,chung2024cfg++}, and extra additional image conditions~\cite{zhao2024unicontrolnet,gal2022image-ti,ruiz2023dreambooth}.
A recent popular line of research focuses on optimizing the initial noise~\cite{guo2024initno,eyring2024reno,qi2024not-all-seed,chen2024find-seed}, motivated by the observation that images generated from the same noise often exhibit high similarities \emph{within a single model}, even when prompted differently. 
In this work, we extend this understanding by revealing that such noise-induced consistency persists across T2I models with \textbf{different diffusion backbones}, as illustrated in \cref{fig:teaser_fig-a}. 

To better understand this phenomenon, we examine the formulation of diffusion models and reveal that cross-model similarities stem from the \emph{finite-step} noise scheduler design and persist across different diffusion variants. Specifically, one of the fundamental assumptions in diffusion models lies in a gradual stepwise perturbation process that transforms a real-world distribution into a standard Gaussian. This transition requires, \emph{in theory}, an infinite number of diffusion steps to perfectly obscure the raw images, contradicting the practical implementation where we can only apply a finite number (i.e., non-asymptotic) of degradations through pre-defined noise schedulers~\cite{li2024towards,chen2023restoration}. As a result, the real-world diffusion models inadvertently learn to leverage residual information embedded in image data, creating an unintended link between noise and data regardless of the model architecture during inference. However, rather than being a limitation, we leverage this seemingly imperfect model design and unintentional connection as an unexpected advantage by laying the foundation for using stochastic noise as a generic ``silent assistant", which implicitly encodes clues about the image that will be generated.



\begin{figure}[t]
    \centering
    \includegraphics[width=0.9\linewidth]{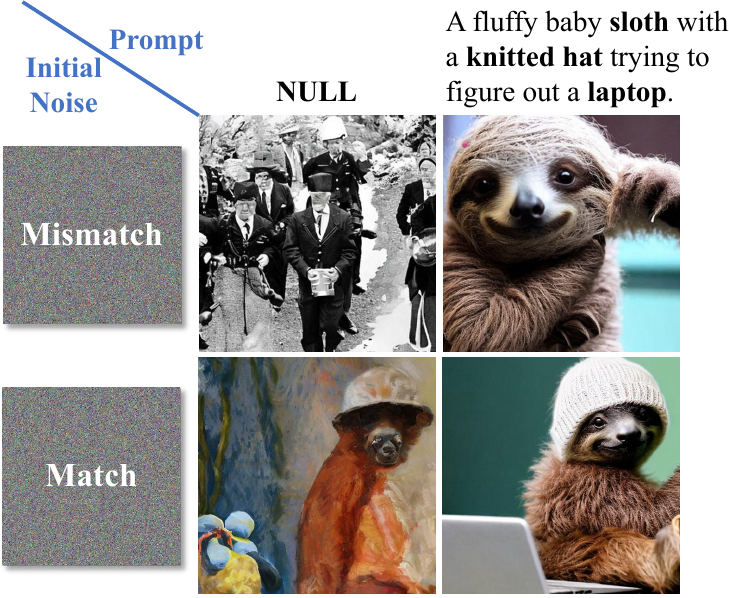}
    \vspace{-1mm}
    \caption{Compared to mismatched noise, noise aligned with text prompts can produce more accurate images in complex T2I generation. A NULL prompt represents unconditional generation.}
    \label{fig:teaser_fig-b}
\end{figure}

We note that it is critical to align the implicit noise clues with the user-provided input, such as text prompts, in order to achieve optimal generation performance. As shown in \cref{fig:teaser_fig-b}, when the inherent tendencies of the noise guidance and text prompt diverge significantly, the generation process leads to instability or semantically inconsistent results. Furthermore, unlike text prompts that convey high-level semantics, initial noise inherently \emph{encodes low-level visual primitives}-such as color gradients, texture patterns, and spatial frequencies—directly tied to the pixel space. This makes it particularly suitable for controlling fine-grained visual properties that are difficult to specify through text alone. By leveraging initial noise as a complementary control signal, we produce more nuanced and precise outputs, enhancing the overall generative capability of T2I models.

Building on this insight, we propose \emph{\textbf{NoiseQuery}}, a novel method that exploits initial noise as a shared generative asset across diverse T2I models. The key challenge lies in establishing an interpretable mapping between stochastic noise patterns and their concrete generative clues. 
To bridge this gap, we utilize images generated with empty NULL text prompts as proxies to reveal the hidden information encoded in the initial noise (details in \cref{sec:initial_noise_analisis}), denoted as the \emph{implicit generative posterior}. When deprived of text guidance, the generation process becomes purely noise-driven, forcing the model to excavate and decode the residual signals preserved in the initial noise. 
This process enables the construction of a \emph{comprehensive Noise Library}, built offline by generating a diverse set of such noise-driven images from a Gaussian distribution and extracting their multi-grained features, spanning high-level semantics and low-level visual attributes. 
During inference, given a specific user request (single or combined goals), we extract its features, perform a quick lookup in the pre-built library, and select the best-matching noise for initialization.
Notably, \emph{NoiseQuery} is \textbf{model-agnostic and serves as
\emph{a reusable resource across multiple diffusion architectures}}.
This ability makes it \textbf{naturally compatible with} diverse enhancement strategies including recent noise optimization approaches, providing a \emph{foundational layer} for more effective generation. Moreover, \emph{NoiseQuery} operates at a \textbf{negligible online computational cost} (0.002 seconds per prompt), ensuring seamless integration into existing workflows while enhancing the controllability and quality of T2I generation.


To validate the effectiveness of \emph{NoiseQuery}, we conduct experiments on low-level property preferences, high-level semantic consistency, and their combinations.
Extensive evaluations with six diffusion-based T2I models~\cite{rombach2022high-ldm,chen2023pixartalpha,sauer2025adversarial-sdturbo}, four types of T2I enhancement methods~\cite{wallace2024diffusion-dpo,eyring2024reno,zhao2024bridging-lavi,chung2024cfg++} and two controllable T2I models~\cite{zhao2024unicontrolnet,gal2022image-ti} show that \emph{NoiseQuery} not only improve the base model’s performance but also complements multiple T2I enhancement methods.
Our approach is also \emph{among the first to address the challenge of subtle low-level preference control with diffusion models}, highlighting a new avenue for fine-grained generative tasks. 
Furthermore, by progressively matching the noise, our method is capable of supporting combinations of generation goals, offering more nuanced control for users.

In summary, our contributions are as follows:
\begin{itemize}
    \item  We reveal the generic generative tendencies within the initial noise and leverage this implicit guidance as a universal generative asset that can be seamlessly integrated with existing T2I models and enhancement methods.
    \item  We introduce \emph{NoiseQuery}, a novel method that retrieves an optimal initial noise from a pre-built large-scale noise library, fulfilling versatile user-specified requirements encompassing both semantic and low-level visual attributes—a relatively under-explored area.
    \item We conduct extensive experiments to demonstrate that our method can improve the general quality of diffusion-based generators and can significantly improve multiple T2I enhancement methods while maintaining a minimal cost and without requiring additional learning.
\end{itemize}

\section{Related Work}
\label{sec:related_work}

\noindent \textbf{Improving Base Text-to-Image Generation.} Recent breakthroughs in text-to-image (T2I) generation have been driven by diffusion-based models, yet these models still suffer from artifacts and misalignment with textual prompts. Current solutions primarily fall into two categories: tuning-based and inference-time methods. Tuning-based methods~\cite{black2023training-rlhf,fan2023dpok,li2025aligning-Diffusion-KTO,yang2024using-D3PO,wallace2024diffusion-dpo} optimize model parameters through techniques like direct preference optimization (DPO)~\cite{wallace2024diffusion-dpo,yang2024using-D3PO,li2025aligning-Diffusion-KTO}. Additionally, LaVi-Bridge~\cite{zhao2024bridging-lavi} enhances textual understanding of T2I models by integrating advanced language models with parameter-efficient adapters. In contrast, inference-time methods~\cite{chung2024cfg++,chefer2023attend,hong2023improving-sag,zhuang2025magnet,li2023divide} dynamically adjust the generation process. For example, CFG++~\cite{chung2024cfg++} improves image quality by addressing non-manifold issues in classifier-free guidance. Our approach complements both strategies by offering improved initialization for generation.


\noindent \textbf{Enhancing T2I with Additional Conditions.}
To address the limitations of text-only conditions in meeting diverse user needs, numerous works~\cite{zhang2023adding-controlnet,chen2023seeing-brain,ruiz2023dreambooth,gal2022image-ti,zhao2024unicontrolnet,cheng2023layoutdiffuse,xiao2024fastcomposer,chen2025subjective-wrq,fan2025rehold,pei2024sowing} have expanded T2I generation through multi-modal conditions. For instance, methods like ControlNet~\cite{zhang2023adding-controlnet} introduce structural signals (e.g., edge and depth maps) to control spatial layouts, while IC-Light~\cite{iclight} adjusts the illumination effects. Meanwhile, customization generation approaches focus on learning user-provided concepts like identity~\cite{gal2022image-ti,ruiz2023dreambooth,ruiz2024hyperdreambooth,wang2024instantid,ye2023ip-adpter}, style~\cite{wang2024instantstyle,frenkel2024implicit-style-lora}, and interaction~\cite{huang2024reversion}.
In this paper, we highlight that the initial noise for diffusion sampling can also serve as a conditional resource. Furthermore, compared to efforts that focus primarily on high-level semantic control, we show that initial noise holds untapped potential for guiding various low-level visual properties (e.g., color, texture, and sharpness).

\noindent \textbf{Initial Noise Optimization for Diffusion Sampling.} 
A growing body of research has focused on optimizing the initial noise in diffusion models, driven by the consensus that it significantly impacts the generated output~\cite{xu2024good-seed,huang2024robin,wangdiffusion-cow,ma2025inferencetimescaling,ahn2024noiseworthguidance}. 
\wry{These methods iteratively refine initial noise through backpropagation at inference, guided by different generative objectives, such as semantic alignment~\cite{eyring2024reno,guo2024initno,qi2024not-all-seed,chen2024find-seed,tang2024inference-DNO}, rare concept generation~\cite{samuel2024generating-rareconcept} and image translation~\cite{greenberg2023s2st}. For instance, ReNO~\cite{eyring2024reno} optimizes noises using various human preference rewards like HPS v2~\cite{wu2023human-hpsv2}, to enhance image quality and align with user intent.}
However, these approaches rely on iterative gradient updates during test-time, which are computationally expensive and prone to numerical instability (e.g., gradient explosion).
In this work, we observe that initial noise exhibits cross-model consistency (Fig.~\ref{fig:teaser_fig-a}), making it an exploitable shared asset. Leveraging this property, we construct a reusable noise library that not only facilitates efficient noise selection but also helps existing noise optimization methods improve generation quality.

\begin{figure}[t]
  \centering
   \includegraphics[width=\linewidth]{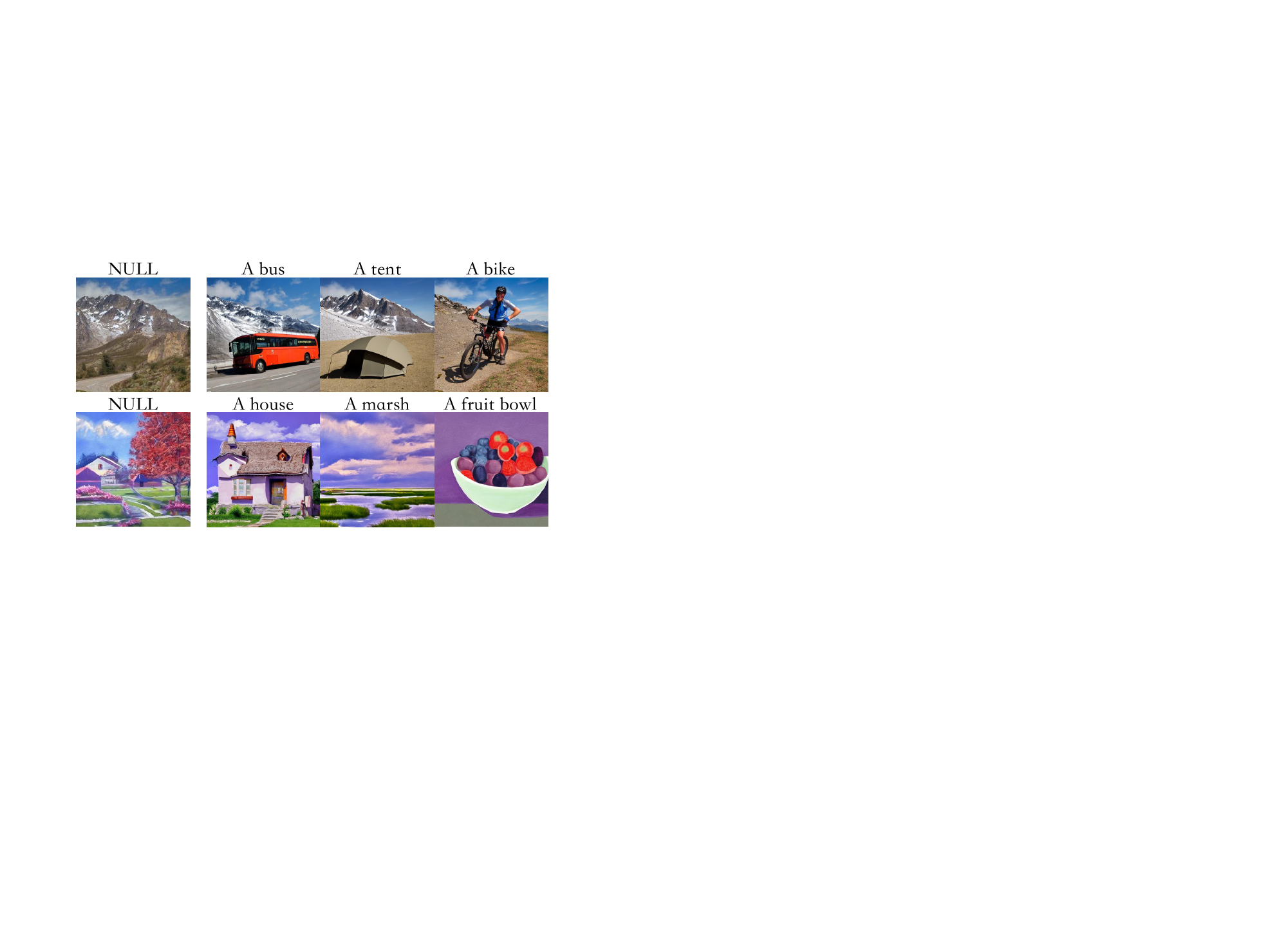}
   \vspace{-5mm}
   \caption{Using the same initial noise, the generative posteriors (left) exhibit similarity with the text-conditioned images (right).
   }
    \vspace{-1mm}
   \label{fig:seed_image}
\end{figure}

\begin{figure*}[t]
  \centering
   \includegraphics[width=0.85\linewidth]{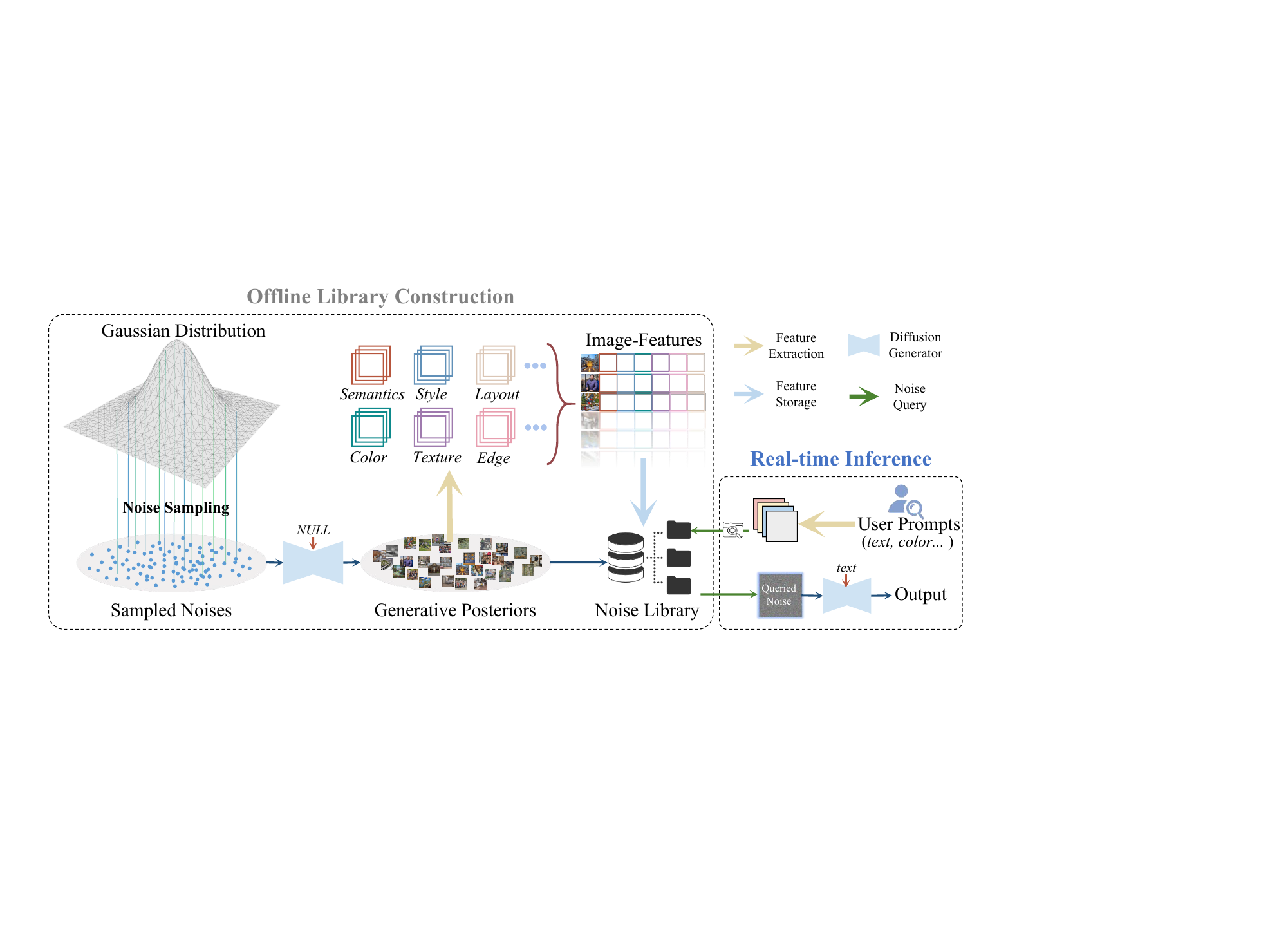}
   \vspace{-2mm}
   \caption{Illustration of our \emph{NoiseQuery} pipeline which involves offline library construction and real-time inference.
   }
   \label{fig:pipeline}
\end{figure*}

\section{Methodology}
\label{sec:method}


\subsection{Problem Formulation}
Given a pre-trained generative diffusion model \( \mathcal{M} \), which is capable of generating images \( \mathcal{I} \) based on user-defined conditions \( c \), typically a text prompt. The generative process starts with the initial noise  \( \epsilon \) sampled from a Gaussian distribution \( \mathcal{N}(0, I) \) and gradually transforms this noise into a clean image through \( T \) denoising steps\footnote{For simplicity, we omit the process of transitioning from pixel space to latent space in latent diffusion models (LDM).}.
Our motivation is to construct a large-scale noise sample library, denoted as \( \mathcal{N}_{set} = \{ \epsilon_1, \epsilon_2, \ldots, \epsilon_n \} \),  where each noise sample \( \epsilon_i \) has its hidden features pre-computed from the generative posteriors. The goal is to identify the most suitable noise \( \epsilon^* \in \mathcal{N}_{set} \) that aligns with a specific generation objective \( \mathcal{O} \). Notably, this generation objective is inherently open-ended and encompasses various dimensions such as semantic consistency, color properties, and other user-defined attributes.

\subsection{Implicit Guidance in Initial Noise}
To investigate the influence of initial noise on generated images, we analyze the forward noise addition process in diffusion models~\cite{ho2020denoising-ddpm,song2020denoising-ddim,song2020score,nichol2021improved-dffusion}. 
The forward diffusion process gradually perturbs an image \( x_0 \) over \( T \) timesteps, transforming it into a ``pure'' Gaussian noise \( x_T \).
The noise addition at each step follows:
\begin{equation}
q( x_t | x_{t-1}) = \mathcal{N}(\sqrt{1-\beta_{t}} x_{t-1}, \beta_{t}\mathbf{I}),
\end{equation}
\noindent 
where \(\beta_t\) is a predefined variances schedule. Defining \(\alpha_t=1-\beta\), the cumulative effect of noise over \( t \) steps results in:
\begin{equation}
x_t = \sqrt{\bar{\alpha}_t} x_0 + \sqrt{1 - \bar{\alpha}_t} \epsilon,
\end{equation}
\noindent
where \( \bar{\alpha}_t = \prod_{i=1}^t \alpha_i \) and \( \epsilon \sim \mathcal{N}(0, I) \). 
A fundamental assumption in diffusion theory is that \( \bar{\alpha}_T\rightarrow0 \) as  \( T\rightarrow\infty \), ensuring that the raw image is fully obscured. However, practical implementations can only apply a finite number of degradations (e.g., \( T = 1000 \)) to approximate this ideal. As analyzed in prior work~\cite{lin2024common-noiseschedule}, common schedules (e.g., Linear~\cite{ho2020denoising-ddpm}, Cosine~\cite{nichol2021improved-dffusion}, Stable Diffusion~\cite{rombach2022high-ldm}) fail to achieve zero signal-to-noise ratio (SNR) at \( T \), leaving residual signal from \( x_0 \) in \( x_T \). For instance, Stable Diffusion’s final timestep retains:
\begin{equation}
x_T = 0.068265 \cdot x_0 + 0.997667 \cdot \epsilon,
\end{equation}
where \( \sqrt{\bar{\alpha}_T}=0.068265 \) quantifies the leakage. During training, diffusion models minimize reconstruction loss by utilizing these residuals as shortcuts rather than predicting the original image from scratch. This creates an unexpected link between noise and data, regardless of the model architecture. As a result, during inference, models continue to rely on the implicit knowledge gained during training to interpret pure Gaussian noise. We take advantage of this seemingly flawed model design and the unintended connection, using stochastic noise as a generic ``silent assistant" that encodes clues about the image to be generated.

\begin{algorithm}[t]
\small
\caption{\emph{NoiseQuery} for Goal-Driven Generation}
\KwIn{
    Diffusion model \( \mathcal{M} \); 
    Noise library \( \mathcal{N}_{set} = \{ \epsilon_1, \epsilon_2, \ldots, \epsilon_n \} \); 
    Generation objective \( \mathcal{O} \) and corresponding matching function \( S(\cdot) \).
}
\KwOut{Optimal noise sample \( \epsilon^* \) for objective \( \mathcal{O} \).}
\vspace{1mm}
\begin{tcolorbox}[colback=gray!10, colframe=white, boxrule=0pt, left=0pt, right=0pt, top=0pt, bottom=0pt]
\textbf{Offline Library Construction} \\
\ForEach{\( \epsilon_i \in \mathcal{N}_{set} \)}{
     \( \mathcal{I}_i^{\text{uncond}} = \mathcal{M}(\epsilon_i, c = \emptyset) \); \\
    Extract features \( \mathcal{F}_i \) (as detailed in Tab.~\ref{Tab:features}) from \( \mathcal{I}_i^{\text{uncond}} \); \\
    Store \( (\epsilon_i, \mathcal{F}_i) \) in the noise library.
}
\end{tcolorbox}

\begin{tcolorbox}[colback=blue!10, colframe=blue!80, boxrule=0.5pt, left=0pt, right=0pt, top=0pt, bottom=0pt]
\textbf{Real-time Inference} \\
Extract desired features \( \mathcal{F}_O \) from objective \( \mathcal{O} \); \\
\( \epsilon^* = \arg\max_{\epsilon_i \in \mathcal{N}_{set}}  \{ S(\mathcal{F}_i, \mathcal{F}_O) \mid \epsilon_i \in \mathcal{N}_{set} \} \) .
\end{tcolorbox}
\vspace{-1mm}
\Return \( \epsilon^* \).
\label{algo:method}
\end{algorithm}

\subsection{Dissect Initial Noise via Generation Posterior}\label{sec:initial_noise_analisis}

While recent works~\cite{guo2024initno,greenberg2023s2st,chen2024find-seed,eyring2024reno,qi2024not-all-seed} optimize initial noise via gradient back-propagation for text-guided generation, they often overlook the inherent generative tendencies embedded in different noises. In other words, if the initial noise deviates significantly from the desired objective, iterative optimization struggles to bridge this gap, resulting in suboptimal convergence despite extensive tuning. In this subsection, we propose a more efficient alternative: directly analyzing the initial noise itself without the text prompt. Additionally, our method also yields better starting points for optimization, leading to faster convergence.


\noindent \textbf{Generative Posterior.}
When the model generates images without any specific prompt, it depends solely on the noise characteristics, producing what we term \textit{``generative posterior''}.
As shown in \cref{fig:seed_image}, generative posteriors exhibit clear generative tendencies (e.g., semantics, spatial layouts, and color palettes) similar to those of text-guided outputs.
These similarities suggest that the initial noise carries significant latent information, influencing both unconditional generation and conditional generation in a similar manner, acting as an additional form of guidance alongside the text prompt. 
Thus, these purely noise-guided images can serve as an effective proxy for interpreting the initial noise.


\noindent \textbf{Cross-model Consistency.}
We also observe another meaningful phenomenon: the impact of initial noise is universal and model-agnostic. As illustrated in \cref{fig:teaser_fig-a}, generative posteriors generated from the same noise remain highly similar across models with different architectures and parameters. For example, PixArt-\(\alpha\)~\cite{chen2023pixartalpha} is a Transformer-based T2I diffusion model, while Stable Diffusion~\cite{rombach2022high-ldm} is UNet-based. More discussion of models and samplers can be found in \cref{sec:sup-sampler}.
This finding empowers us to create a large-scale, model-agnostic noise library, where each noise sample can be associated with diverse latent features and then shared across different models. 
The universality of noise-driven priors positions initial noise as a unified control resource for various applications.

\begin{table}[t]
\centering
\resizebox{.5\textwidth}{!}{
\begin{tabular}{lcc}
\toprule
\textbf{Generation Goals} & \textbf{Feature Type} & \textbf{Match Function} \\
\midrule
\textcolor{highLevelColor}{\textbf{\textit{Semantics}}} & CLIP~\cite{radford2021learning-clip}, BLIP~\cite{li2022blip} & Cosine Similarity \\
\textcolor{highLevelColor}{\textbf{\textit{Style}}} & Gram Matrix~\cite{gatys2015neural-style} & MSE \\
\hline
\textcolor{lowLevelColor}{\textbf{\textit{Color}}} & RGB, HSV, LAB & Absolute Difference \\
\textcolor{lowLevelColor}{\textbf{\textit{Texture}}} & GLCM~\cite{haralick1973textural-glcm} & Euclidean Distance \\
\textcolor{lowLevelColor}{\textbf{\textit{Shape}}} & Hu Moments~\cite{hu1962visual-hu} & Euclidean Distance \\
\textcolor{lowLevelColor}{\textbf{\textit{Sharpness}}} & High Frequency Energy (HFE) & Absolute Difference \\
\bottomrule
\end{tabular}
}
\vspace{-2mm}
\caption{Feature types and matching functions.
}
\label{Tab:features}
\end{table}

\subsection{NoiseQuery for Goal-Driven Generation}
Building upon the insights regarding initial noise characteristics and cross-model consistency, we propose \emph{NoiseQuery}, an efficient approach for selecting appropriate noise samples tailored to diverse generation objectives. As illustrated in \cref{fig:pipeline}, our method consists of two main phases: offline library construction and real-time inference. We also provide pseudo-code in \cref{algo:method}.

\noindent \textbf{Offline Library Construction.} 
In the first stage, we build a noise library \( \mathcal{N}_{set} \), which contains many initial noise samples (e.g., 100k) sampled from standard Gaussian distribution.
These samples are utilized to generate generative posteriors by running a diffusion model without any user-defined prompts.
Subsequently, we extract a variety of features \( \mathcal{F}_i \) from these generative posteriors that correspond to various generative objectives  \( \mathcal{O} \), such as semantic representations (e.g., CLIP features), color properties (e.g., saturation),  etc. 
These extracted features are then stored as keys alongside their corresponding noise samples (values) within the noise library, creating a comprehensive key-value repository that can be queried during the inference phase.

\noindent \textbf{Real-time Inference.} 
When performing real-time inference, we specify desired features \( \mathcal{F}_O \) following the user's generation objective \( \mathcal{O} \). 
These features could be semantic, color, spatial, or style-based attributes, depending on the specific goal. 
Using a match function \( S(\cdot) \), we evaluate the alignment between each feature \( \mathcal{F}_i \) in our pre-computed library and the objective features \( \mathcal{F}_O \). 
\wry{The optimal noise sample \( \epsilon^* \) is selected by maximizing the matching scores:}
\vspace{-1mm}
\begin{equation}
\epsilon^* = \arg\max_{\epsilon_i \in \mathcal{N}_{set}} \{ S(\mathcal{F}_i, \mathcal{F}_O) \mid \epsilon_i \in \mathcal{N}_{set} \}.
\vspace{-1mm}
\end{equation}
Once the optimal noise sample  \( \epsilon^* \) is identified, it is fed into the generative model to produce an image that closely matches the user intent. 
\wry{In \cref{Tab:features}, we summarize several common generation goals achievable with \emph{NoiseQuery}, including both high-level semantics and low-level visual properties. For each goal, we specify the corresponding feature types and matching functions. Detailed explanations of these features can be found in \cref{sec:sup-lowlevel}. 
Notably, these goals can be applied individually or in combination. Users can specify multiple constraints simultaneously (e.g., generating a ``sunset” image with a ``warm color palette” while ensuring ``high sharpness”). This flexibility makes \emph{NoiseQuery} highly versatile and generalizable, enabling fine-grained control over the generated output.
}

\section{Experiments}
\label{sec:experiment}

\begin{table*}[]
\centering
\normalsize
\resizebox{\textwidth}{!}{
\begin{tabular}{llcccccccccc}
\toprule
\multirow{2}{*}{\textbf{Base Model}} & \multirow{2}{*}{\textbf{Method}}                                        &\multicolumn{4}{c}{\textbf{DrawBench~\cite{saharia2022photorealistic-drawbench}}}                       & \multicolumn{4}{c}{\textbf{MSCOCO~\cite{lin2014microsoft-coco}}}   &   \multirow{2}{*}{\textbf{Time Cost}}                                                                                  \\ 
\cmidrule(lr){3-6} \cmidrule(lr){7-10}
                                      &                                            &                                 \textbf{ImageReward} & \textbf{PickScore}   & \textbf{HPS v2} & \textbf{CLIPScore} & 
                                      \multicolumn{1}{c}{\textbf{ImageReward}} & \multicolumn{1}{c}{\textbf{PickScore}} & \textbf{HPS v2} & \textbf{CLIPScore} & 
                                      \\ 
                                      \midrule
\multirow{4}{*}{SD 1.5} & Base Model  & 0.04 & 21.11& 24.57 &30.90    & 0.15 & 21.41 & 25.65 & 31.08 & 1.334 s \\
 &~~+ \textcolor{noisequerycolor}{NoiseQuery} & \textbf{0.08} &\textbf{21.16}& \textbf{25.02} & \textbf{31.41}  & \textbf{0.27} & \textbf{21.48} & \textbf{26.07}&\textbf{31.47}  & 1.336 s \\
 \noalign{\vspace{0.8mm}}
                                      \cdashline{2-11} 
                                      \noalign{\vspace{0.8mm}}
 &~~+ \textcolor{t2ienhance}{Diffusion-DPO~\cite{wallace2024diffusion-dpo}}  & 0.09 & 21.29 & 25.02 & 31.19  & 0.25 & 21.64 & 26.31 & 31.26 & 1.350 s \\
 &~~+ \textcolor{t2ienhance}{Diffusion-DPO~\cite{wallace2024diffusion-dpo}} + \textcolor{noisequerycolor}{NoiseQuery} & \textbf{0.17} & \textbf{21.33} &  \textbf{25.25}&\textbf{31.41}  & \textbf{0.35 }& \textbf{21.68} & \textbf{26.60} & \textbf{31.55 }& 1.352 s \\                      \midrule
\multirow{4}{*}{SD 2.1}& Base Model & 0.12  & 21.33  & 24.93  & 31.13   &   0.36   & 21.72   &    26.58      &   31.40     &  1.301 s  &  \\
                                      &~~+ \textcolor{noisequerycolor}{NoiseQuery}                            & \textbf{0.26 }                & \textbf{21.46}       & \textbf{25.39}           & \textbf{31.68}   & \textbf{0.44}&\textbf{21.76}&\textbf{26.82}&   \textbf{31.50}                 &      1.303 s              \\
                                      \noalign{\vspace{0.8mm}}
                                      \cdashline{2-11} 
                                      \noalign{\vspace{0.8mm}}
                                      &~~+ \textcolor{t2ienhance}{CFG++~\cite{chung2024cfg++}}                                       & 0.12                 & 21.33                & 24.83           & 31.13        & 0.37                                         & 21.72 &    26.66     & 31.31 &  3.724 s                              \\
                                       &~~+ \textcolor{t2ienhance}{CFG++~\cite{chung2024cfg++}} + \textcolor{noisequerycolor}{NoiseQuery}                                          & \textbf{0.27}    & \textbf{21.43}                & \textbf{25.55}  & \textbf{31.61}   &\textbf{0.47} & \textbf{21.76} & \textbf{26.97} &   \textbf{31.67}     & 3.726 s                      \\
                                      \midrule
\multirow{4}{*}{SD-Turbo}      & Base Model                  &   0.26                 & 21.78                & 25.23           & 31.29   &0.47&22.07& 26.22& 31.51& 0.072 s\\

                                      &~~+ \textcolor{noisequerycolor}{NoiseQuery}                           & \textbf{0.41} & \textbf{21.87} &    \textbf{25.66 }            & \textbf{31.58}              &   \textbf{0.50}   &\textbf{22.17}&\textbf{26.82}&\textbf{31.76}& 0.074 s                    \\ 
                                      \noalign{\vspace{0.8mm}}
                                      \cdashline{2-11} 
                                      \noalign{\vspace{0.8mm}}
                                      &~~+ \textcolor{t2ienhance}{ReNO~\cite{eyring2024reno}}                              & 1.67                 & 23.40                & 32.48           & 32.55                &-&-&-&-&      23.56 s              \\  
                                      &~~+ \textcolor{t2ienhance}{ReNO~\cite{eyring2024reno}} + \textcolor{noisequerycolor}{NoiseQuery}                                          & \textbf{1.71}        & \textbf{23.52}       & \textbf{32.92}  & \textbf{32.78}                     &-&-&-&-&   23.56 s                 \\ 
                                      \midrule
\multirow{4}{*}{PixArt-$\alpha$}      & Base Model                  &0.70&22.08&28.27&30.83&0.78&22.24&29.33&31.48 & 4.327 s\\

                                      &~~+ \textcolor{noisequerycolor}{NoiseQuery}
                                      & \textbf{0.82} &\textbf{22.11}&\textbf{28.45}&\textbf{31.27}&  \textbf{0.79}&\textbf{22.33}&\textbf{29.56}&\textbf{31.64}&  4.328 s                 \\  \noalign{\vspace{0.8mm}}
                                      \cdashline{2-11} 
                                      \noalign{\vspace{0.8mm}}
                                      &~~+ \textcolor{t2ienhance}{LaVi-Bridge~\cite{zhao2024bridging-lavi}}                              &0.63&22.08&28.35&30.92&0.75&22.31&29.49&31.86&  5.092 s                  \\
                                      &~~+ \textcolor{t2ienhance}{LaVi-Bridge~\cite{zhao2024bridging-lavi}} + \textcolor{noisequerycolor}{NoiseQuery}
                                      &\textbf{0.72}&\textbf{22.24}&\textbf{28.61}&\textbf{31.35}&\textbf{0.78}&\textbf{22.35}&\textbf{29.67}&\textbf{32.01}& 5.094 s                \\ 
                                      \bottomrule                  
\end{tabular}
}
\caption{Evaluation of objective metrics on different datasets and models. 
Higher is better for all metrics. \textcolor{noisequerycolor}{Our approach} enhances the base model's performance and complements a wide range of \textcolor{t2ienhance}{T2I enhancement methods}, including \hhy{DPO~\cite{wallace2024diffusion-dpo} (reward-based fine-tuning), CFG++~\cite{chung2024cfg++} (guidance refinement), ReNO~\cite{eyring2024reno} (gradient-based initial noise optimization), and LaVi-Bridge~\cite{zhao2024bridging-lavi} (text encoding enhancement via LORA/adapters).} `-': ReNO results on MSCOCO are excluded due to excessive inference time. 
}
\vspace{-3mm}
\label{Tab:semantic-benchmark}
\end{table*}

\subsection{Experimental Setup}

\noindent \textbf{Datasets.} We evaluate our approach on two popular benchmarks: MSCOCO~\cite{lin2014microsoft-coco} and DrawBench~\cite{saharia2022photorealistic-drawbench}. MSCOCO is a standard benchmark for text-to-image generation, providing textual descriptions of diverse natural scenes and object classes. We randomly select 10k captions as user prompts. DrawBench consists of 200 carefully designed prompts intended to challenge generative scenarios on various aspects, including long-form text, complex compositions, conflicting instructions, and rare vocabulary.

\noindent \textbf{Model Zoo.}
We conduct experiments on six popular diffusion generation models, including Stable diffusion (SD) series~\cite{rombach2022high-ldm} (1.4, 1.5, 2.0, 2.1),  PixArt-\(\alpha\)~\cite{chen2023pixartalpha}, and SD-Turbo~\cite{sauer2025adversarial-sdturbo}. 
Stable Diffusion is a Latent Diffusion Model with UNet~\cite{ronneberger2015u-unet} backbone. By default, we use 50 DDIM denoising steps with a CFG scale of 7.0. PixArt-\(\alpha\) is a Transformer Latent Diffusion Model, and SD-Turbo is a distillation model based on Stable Diffusion 2.1, generating high-quality images even with only 1-2 sampling steps without CFG. For PixArt-\(\alpha\), we use the default CFG scale of 4.5.

\noindent \textbf{Evaluation Metrics.}
For semantic similarity, we use ClipScore~\cite{radford2021learning-clip}, which directly measures the alignment between the generated images and the corresponding textual prompts. Additionally, we incorporate several popular metrics trained on large-scale human preference data to quantify human preferences for AI-generated images, including HPS v2~\cite{wu2023human-hpsv2}, PickScore~\cite{kirstain2023pick-pickscore}, and ImageReward~\cite{xu2024imagereward}. These metrics offer a thorough evaluation of image quality, aesthetics, and alignment with user intent. 

\begin{figure}[t]
  \centering
   \includegraphics[width=\linewidth]{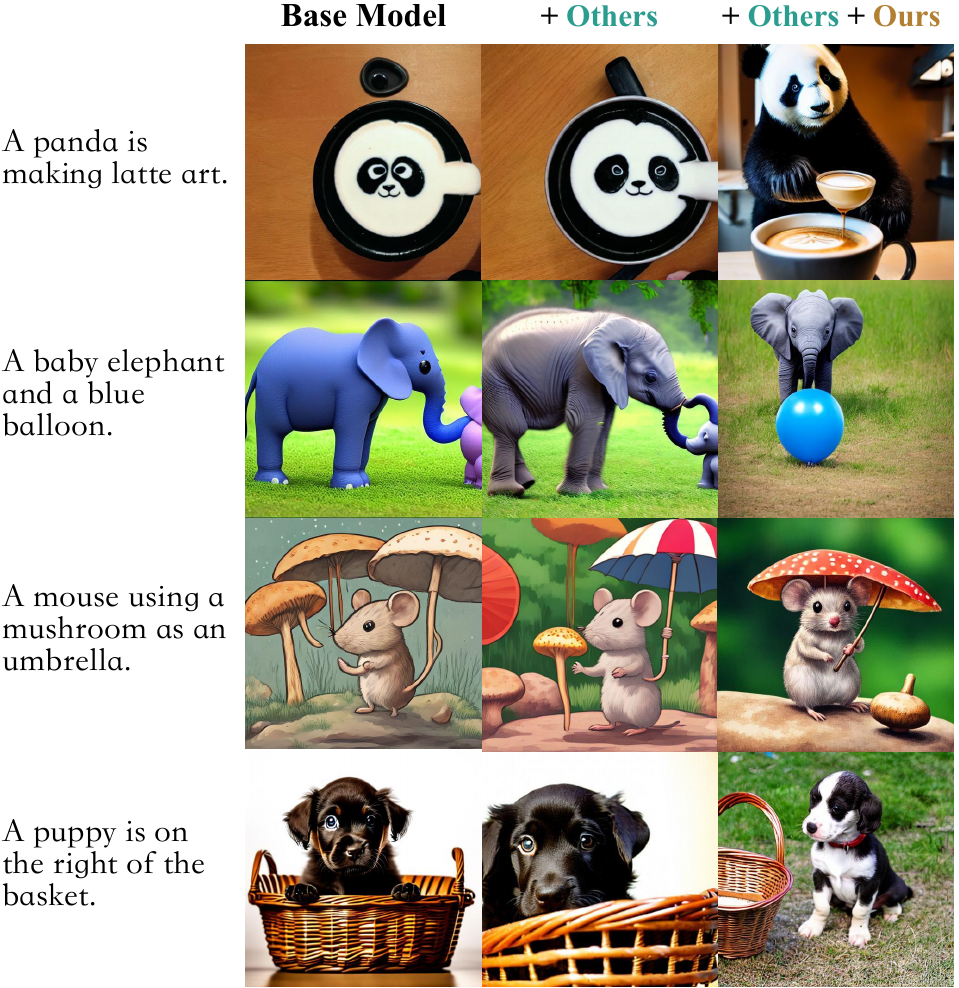}
   \vspace{-5mm}
   \caption{
Generated images of the base model, base model with \textcolor{t2ienhance}{T2I enhancement methods}, and base model with \textcolor{t2ienhance}{enhancement} and \textcolor{noisequerycolor}{\textit{NoiseQuery}}. The four rows, from top to bottom, respectively correspond to the method groups listed in Table~\ref{Tab:semantic-benchmark}: Diffusion-DPO~\cite{wallace2024diffusion-dpo}, CFG++~\cite{chung2024cfg++}, ReNO~\cite{eyring2024reno}, and LaVi-Bridge~\cite{zhao2024bridging-lavi}.
   }
    \vspace{-3mm}
   \label{fig:comp_aug}
\end{figure}

\subsection{Improvement on High-level Semantics}\label{sec:exp-semantics}

For semantic consistency goals, we utilize features like CLIP~\cite{radford2021learning-clip} and BLIP~\cite{li2022blip} as the query to identify the semantically matched noise for each text prompt. Such aligned text and noise yield two benefits: (i) prevents the model from generating irrelevant outputs or falling into unsolvable regions. (ii) minimizes generation difficulty and maximizes the likelihood for semantic consistency.


\begin{figure}
    \centering
        \includegraphics[width=.9\linewidth]{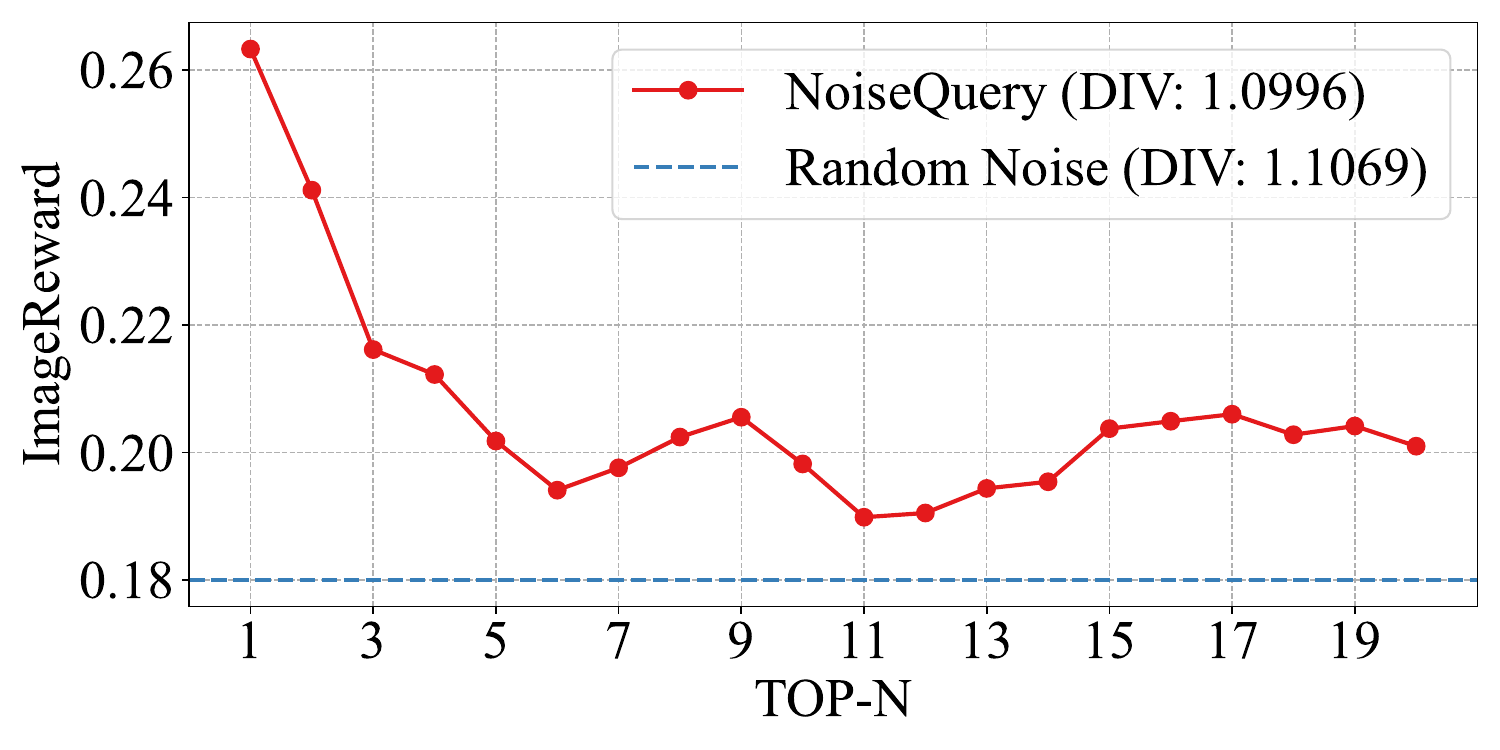}
        \vspace{-3mm}
        \captionof{figure}{Queried noise v.s. random noise.
        }
        \vspace{-3mm}
        \label{fig:top20}
\end{figure}

\begin{figure}[t]
    \centering
    \includegraphics[width=.9\linewidth]{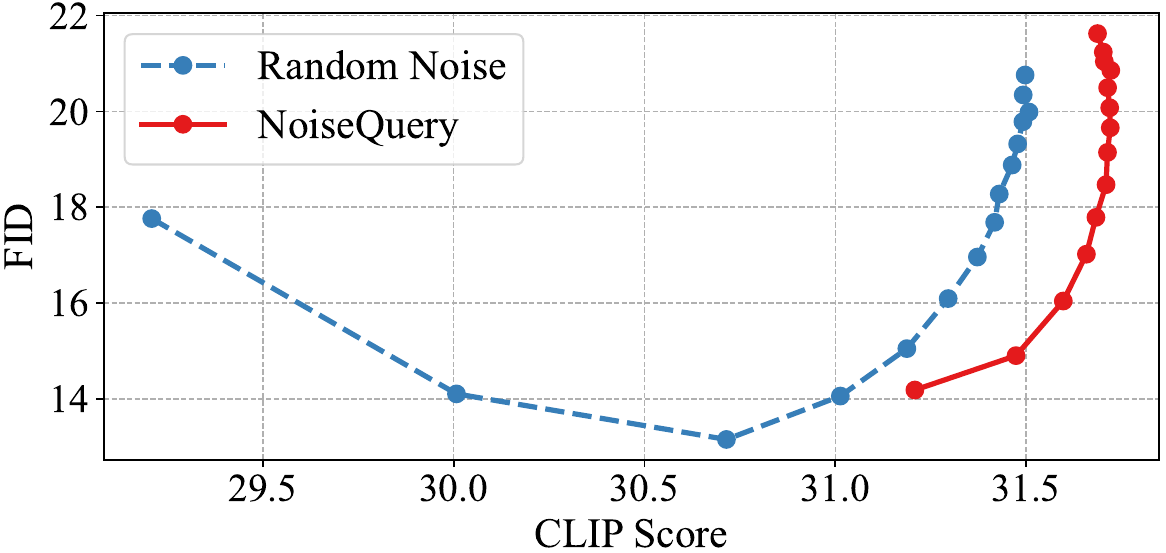}
    \vspace{-3mm}
    \caption{
    CLIP and FID evaluations across CFG scales on MSCOCO~\cite{lin2014microsoft-coco}, with the CFG scale increasing from left to right.
    }
    \label{fig:coco_curve}
    \vspace{-5mm}
\end{figure}

\begin{figure*}[t]
    \centering
    \begin{subfigure}[t]{0.48\textwidth}
        \centering
    \includegraphics[width=\linewidth]{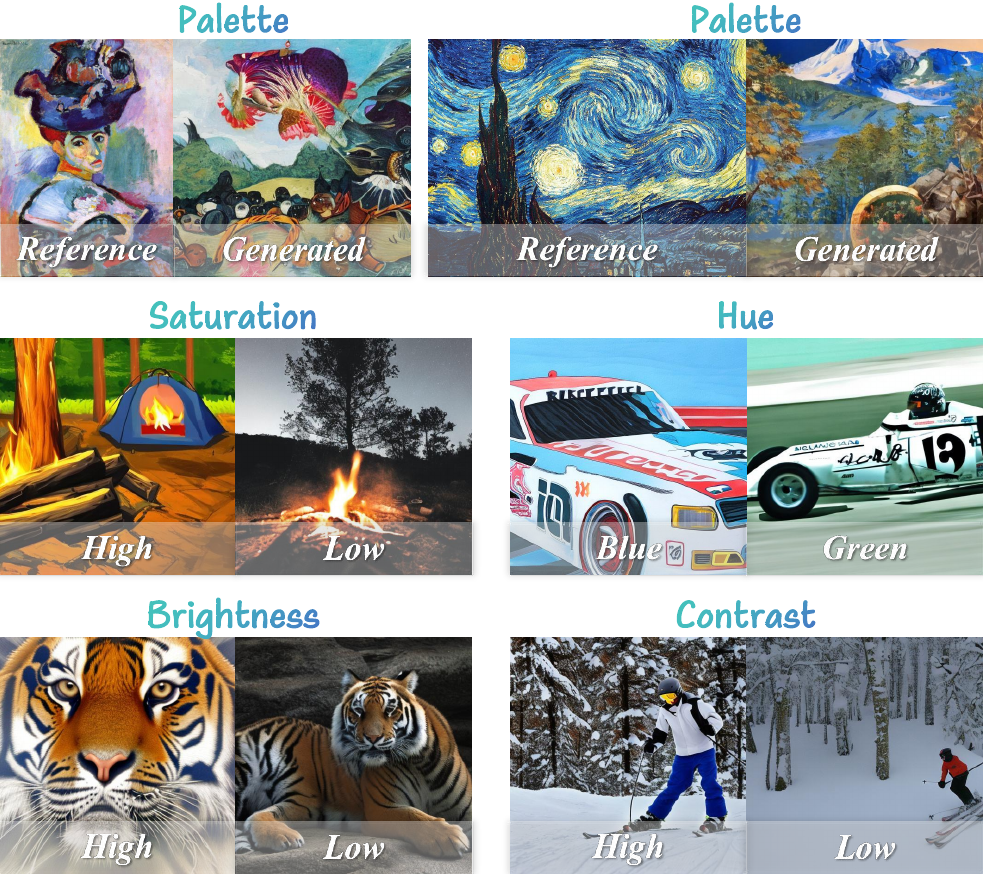}
        \caption{Color properties guidance.}
        \label{fig:color_attribute}
    \end{subfigure}
    \hfill
    \begin{subfigure}[t]{0.48\textwidth}
        \centering
    \includegraphics[width=\linewidth]{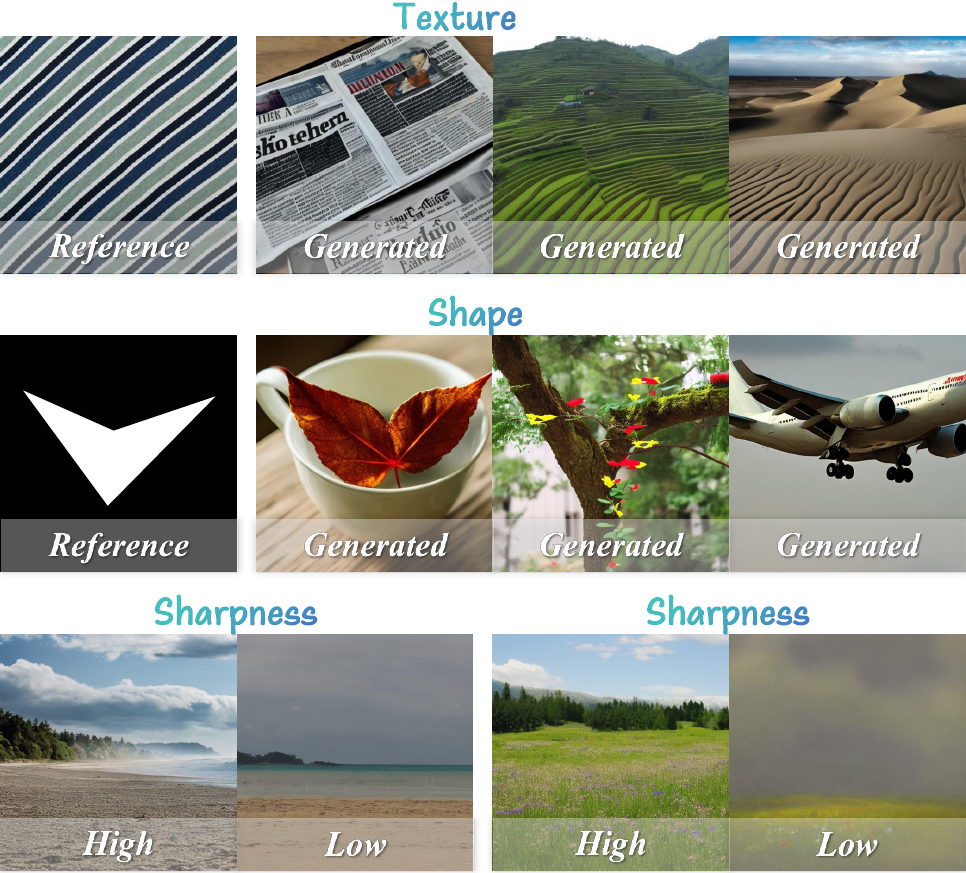}
        \caption{Structural features guidance.}
        \label{fig:lowlevel}
    \end{subfigure}
    \vspace{-3mm}
    \caption{Multiple low-level visual properties control through \emph{NoiseQuery}. \textbf{All images use minimal text prompts containing only semantic concepts (e.g., object labels), lacking explicit visual specifications}. 
    }
    \label{fig:all-lowlevel}
\end{figure*}
\noindent \textbf{Quantitative Analysis.}
In \cref{Tab:semantic-benchmark}, we evaluate our method on standard text-to-image (T2I) benchmarks, demonstrating consistent performance gains across all base models by offering better noise-text alignment. \textit{NoiseQuery} also serves as a universal foundation layer, complementing T2I enhancement techniques like Diffusion-DPO~\cite{wallace2024diffusion-dpo}, CFG++~\cite{chung2024cfg++}, ReNO~\cite{eyring2024reno}, and LaVi-Bridge~\cite{zhao2024bridging-lavi}.
Notably, these methods face inherent limitations: model-specific dependencies (e.g., Diffusion-DPO and LaVi-Bridge require retraining for new models; ReNO is restricted to one-step models due to heavy computation, excluding common multi-step models like SD) and high latency. In contrast, our method seamlessly integrates with them, enhancing their performance while adding near-zero time costs.
We also present qualitative results in \cref{fig:comp_aug} to illustrate its effectiveness in assisting text-to-image synthesis.
\wry{Additionally, we show the zero-shot transferability of \textit{NoiseQuery} across various generative models in \cref{sec:sup-zero_shot_transfer}.}
This model-agnostic compatibility, combined with minimal computational cost, establishes \textit{NoiseQuery} as a versatile and efficient enhancement tool for diverse pipelines.



\begin{table}[t]
\footnotesize
\centering
\setlength{\tabcolsep}{3pt} 
\resizebox{\linewidth}{!}{
\begin{tabular}{lcccccccc}
\toprule
\textbf{Library Size} & \textbf{0.5k} & \textbf{1k} & \textbf{2k} & \textbf{5k} & \textbf{10k} & \textbf{50k} & \textbf{100k}\\ 
\midrule
\textbf{Matching Function Cost (\boldmath$\times 10^{-4}$ s)}& 1.39 & 1.39 & 1.39 & 1.39 & 1.40 & 1.40 & 1.51 \\ 
\textbf{\wry{Argmax Selection Cost} (\boldmath$\times 10^{-4}$ s)} & 0.25 & 0.25 & 0.25 & 0.25 & 0.37 & 6.16 & 13.25  \\ 
\textbf{CLIP Score} & 31.51 & 31.53 & 31.57 & 31.59 & 31.66 & 31.73 & 31.74 \\ 
\bottomrule
\end{tabular}}
\vspace{-2mm}
\caption{Retrieval time breakdown across various library sizes. }
\label{tab:retrieval_time_compact}
\end{table}

\noindent \textbf{Diversity Analysis.}
We report the performance of selecting the top 20 matched noise samples per prompt instead of the single best match.
As shown in \cref{fig:top20}, these samples consistently outperform the standard random noise baseline on DrawBench~\cite{saharia2022photorealistic-drawbench}, demonstrating that \textit{NoiseQuery} provides not only optimal but also diverse high-quality noise candidates.
To quantify diversity, we use the Diversity Metric (DIV)~\cite{wang2024analysis}, computed as the standard deviation of DINOv2~\cite{oquab2023dinov2} embeddings across 20 generated images per prompt (see \cref{fig:top20} legend).
Our method achieves comparable diversity to random noise, confirming that \textit{NoiseQuery} enhances image quality without sacrificing diversity.

\noindent \textbf{Different Classifier-free Guidance Scale Analysis.}
We assess our method's performance across various classifier-free guidance (CFG) scales (1.5, 2.0, 3.0, ..., 15.0) on 10k MSCOCO~\cite{lin2014microsoft-coco} samples. Our \emph{NoiseQuery} achieves high-quality outputs even with low guidance scales, as demonstrated in \cref{fig:coco_curve}. This supports our hypothesis that by selecting more compatible initial noise, we can effectively reduce the generation difficulty. Therefore, we can obtain high-quality output at low scales and avoid the over-saturation and instability that usually come with high scales.


\noindent \textbf{Time Cost.}
We also provide a cost breakdown for the retrieval process in \cref{tab:retrieval_time_compact}, which includes the computation of the matching function (utilizing query CLIP features at no extra cost since CLIP is embedded in SD) and \wry{argmax selection}.
The results demonstrate that both steps are efficient, even with large libraries (e.g., 0.0015 seconds for 100,000 entries), indicating that the additional cost is negligible.



\begin{figure}[t]
    \centering
    \vspace{-3mm}
    \includegraphics[width=\linewidth]{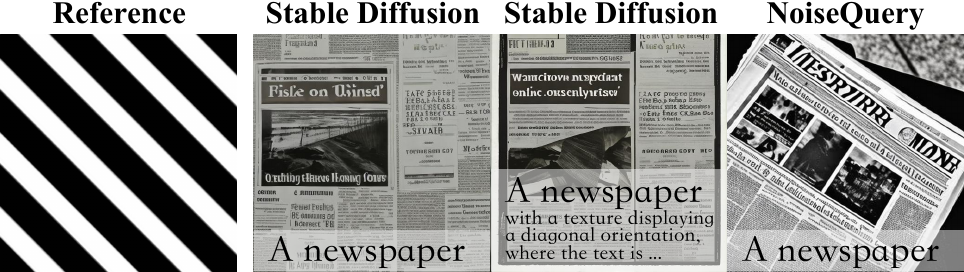}
    \vspace{-6mm}
    \caption{Text prompts often fail to align generated images with user low-level attribute preferences, while initial noise plays a dominant role in controlling them.}
    \label{fig:noise-vs-prompt}
\end{figure}

\subsection{Controllability on Low-Level Visual Properties}\label{sec:exp-lowlwvwl}
Text prompts effectively steer high-level semantics in text-to-image (T2I) generation but often fail to provide sufficient control over low-level visual properties that affect visual perception.
This limitation stems from two factors: (1) text is inherently inadequate for conveying detailed visual reference input, and (2) text encoders like CLIP~\cite{radford2021learning-clip}, trained primarily on high-level semantics, struggle to capture and control fine-grained low-level properties such as color, texture, shape, and sharpness.

In contrast, initial noise plays a dominant role in determining these properties, making it a valuable resource for precise control, as illustrated in \cref{fig:noise-vs-prompt}. Our method leverages this by enabling fine-grained control of low-level attributes through initial noise, without requiring additional training. The following experiments demonstrate the effectiveness of our approach in achieving these targets, with further implementation details provided in \cref{sec:sup-lowlevel}.

\noindent \textbf{Color Properties.}
Color attributes such as brightness, saturation, contrast, and hue are crucial for an image's visual appeal. We assess the overall color hue by averaging each RGB channel, measure saturation and brightness in the HSV color space, and compute contrast as the standard deviation of grayscale intensities. As shown in \cref{fig:color_attribute}, selecting initial noise that matches target color distributions enables precise control over the generated images' color tendencies, aligning them with user preferences.

\begin{figure}[t]
    \centering
    \includegraphics[width=\linewidth]{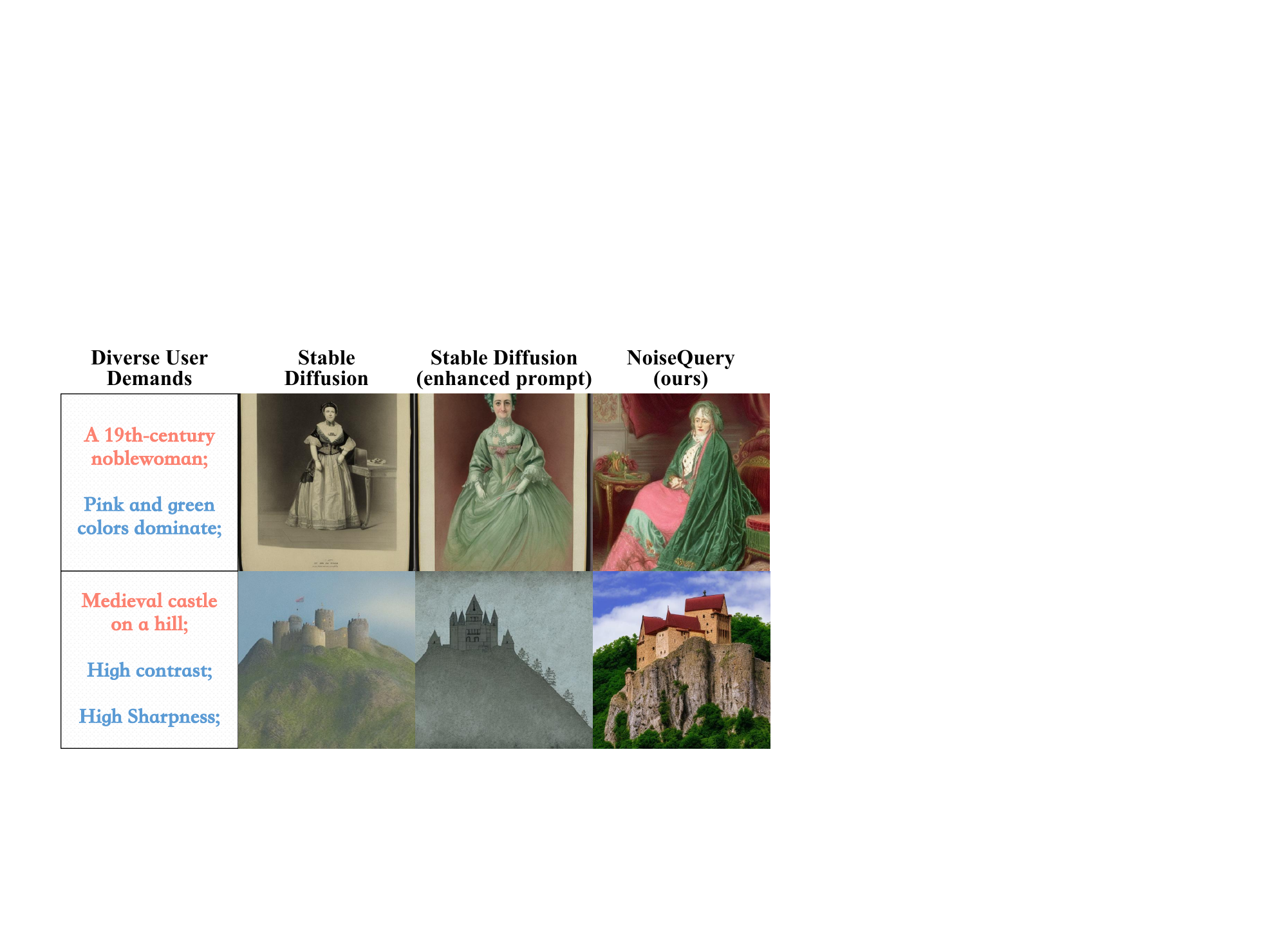}
    \vspace{-5mm}
    \caption{Examples of combining high-level semantics (red) and low-level properties (blue).
    Enhanced prompts are created by fusing semantic objects and visual details via a large language model.
    }
    \label{fig:combinegoals}
\end{figure}

\begin{figure}[t]
    \centering
    \includegraphics[width=\linewidth]{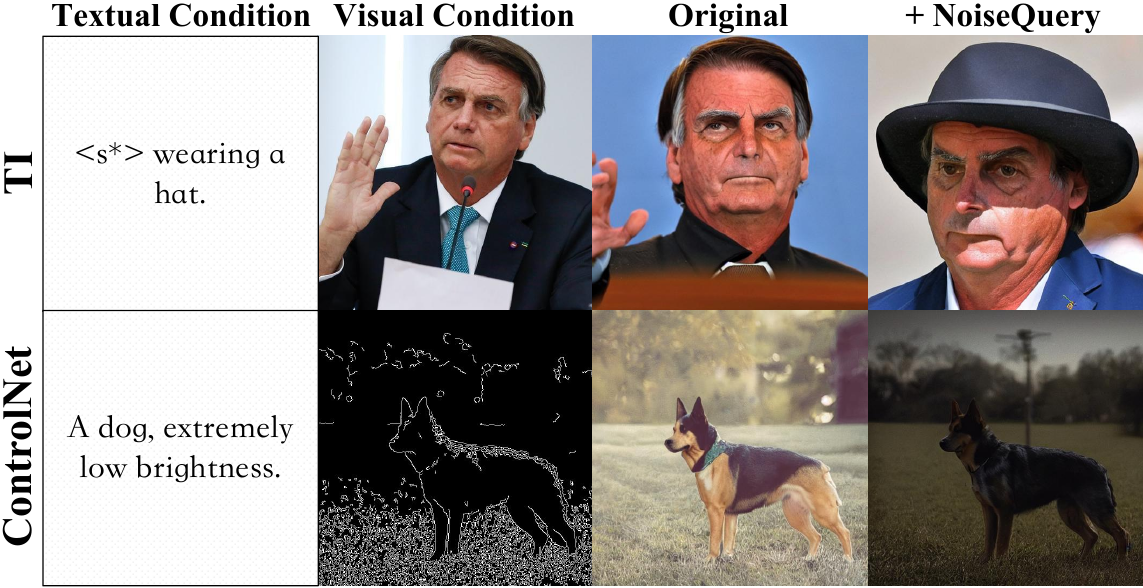}
    \vspace{-5mm}
    \caption{Compatibility with controllable models such as TI~\cite{gal2022image-ti} and ControlNet~\cite{zhang2023adding-controlnet}. Our method provides a better initial noise, enhancing semantic consistency and low-level attribute control. }
    \label{fig:controllabel_models}
\end{figure}

\noindent \textbf{Structural Features.}
For spatial control, we implement: (1) \textit{Texture} through Gray-Level Co-occurrence Matrix (GLCM)~\cite{haralick1973textural-glcm} features that quantify intensity co-occurrence patterns; (2) \textit{Shape} consistency via Hu Moments~\cite{hu1962visual-hu} as rotation-invariant descriptors; and (3) \textit{Sharpness} by prioritizing high-frequency energy (HFE) in Fourier spectra. \cref{fig:lowlevel} demonstrates how noise selection governs these structural features—preserving geometric fidelity in object shapes, modulating surface textures, and enhancing edge crispness—all achieved without prompt engineering.

\subsection{Enhancement on Combined Scenarios}\label{sec:exp-combine}
Building upon the results in previous sections, which demonstrate our method's effectiveness in high-level semantic alignment and low-level attribute control, this section highlights its ability to \emph{combine these diverse goals and seamlessly integrate with powerful existing controllable generation models}.

\noindent \textbf{Combined Goals.}
Beyond individual tasks, our method excels in simultaneously achieving multiple generation goals. By sequentially reranking noise candidates—first selecting top candidates based on one goal (e.g., high-level semantics) and then refining within this subset for another goal (e.g., low-level attributes)—we progressively narrow down the optimal noise pool. As shown in \cref{fig:combinegoals}, this approach allows the combination of high-level semantic alignment with various low-level attribute controls, enabling the generation of images that meet diverse user preferences.

\noindent \textbf{Compatibility with Controlling Methods.}
Our method is designed to be instant-time and model-agnostic, making it highly compatible with existing controllable generation models such as Textual Inversion (TI)~\cite{gal2022image-ti} and ControlNet~\cite{zhang2023adding-controlnet}. By providing a better initial noise, our approach establishes a foundational layer for controllable generation, which synergizes with subsequent refinement processes. As shown in \cref{fig:controllabel_models}, this integration enhances semantic consistency while enabling precise control over challenging low-level attributes, such as color and texture.

\section{Conclusion}
\label{sec:conclusion}


\noindent In this work, we propose \emph{NoiseQuery}, a \textbf{tuning-free} method that enables versatile goal-driven T2I generation through enhanced noise initialization, supported by a generic and reusable pre-built noise library. By selecting Gaussian initialization in the noise library better aligned with the text prompt or specified attributes, \emph{NoiseQuery} achieves fine-grained control over both high-level semantic and intricate low-level details. 
In addition, the model-agnostic nature of our method allows it to be seamlessly integrated with existing T2I augmentation and controllable generation methods without model adjustments or significant computational overhead. Experiments validate its effectiveness in improving output fidelity and precision, particularly for visual attributes that are challenging to articulate textually. The lightweight and scalability of the framework make it adaptable to real-world applications.

\noindent \textbf{Limitations.}
Although \emph{NoiseQuery} provides flexibility for aligning high-level semantics and controlling low-level attributes, it remains bounded by the base model’s ability, and may still underperform in extremely complex cases. Additionally, it has limitations in achieving fine-grained control, such as Canny edge control.

\section*{Acknowledgements}

This work was partially supported by the National Natural Science Foundation of China under grant 62372341. This work was also partially supported by an Amazon Research Award 2023 to OR and YZ. Any opinions, findings, and conclusions or recommendations expressed in this material are those of the authors and do not reflect the views of Amazon.
{
    \small
    \bibliographystyle{ieeenat_fullname}
    \bibliography{main}
}
\addtocontents{toc}{\protect\setcounter{tocdepth}{0}}

\appendix


\clearpage
\setcounter{page}{1}
\setcounter{table}{0}
\setcounter{figure}{0}
\setcounter{algocf}{0}
\maketitlesupplementary

\addtocontents{toc}{\protect\setcounter{tocdepth}{2}}
\tableofcontents

\renewcommand{\thetable}{S\arabic{table}}
\renewcommand{\thefigure}{S\arabic{figure}}

\section*{Overview} 

In \cref{sec:sup-sampler}, we introduce the generative posterior concept, emphasizing its consistency across different models and samplers.
Moreover, \cref{sec:sup-more_semantics} provides a detailed analysis of high-level semantic generation results, exploring our zero-shot transferability, complex image generation, and the effects of CFG scales. Lastly, implementation details for experiments on low-level visual properties are outlined in \cref{sec:sup-lowlevel}, along with the noise offset method for color tendency control.

\begin{figure*}[t]
    \centering
    \includegraphics[width=\linewidth]{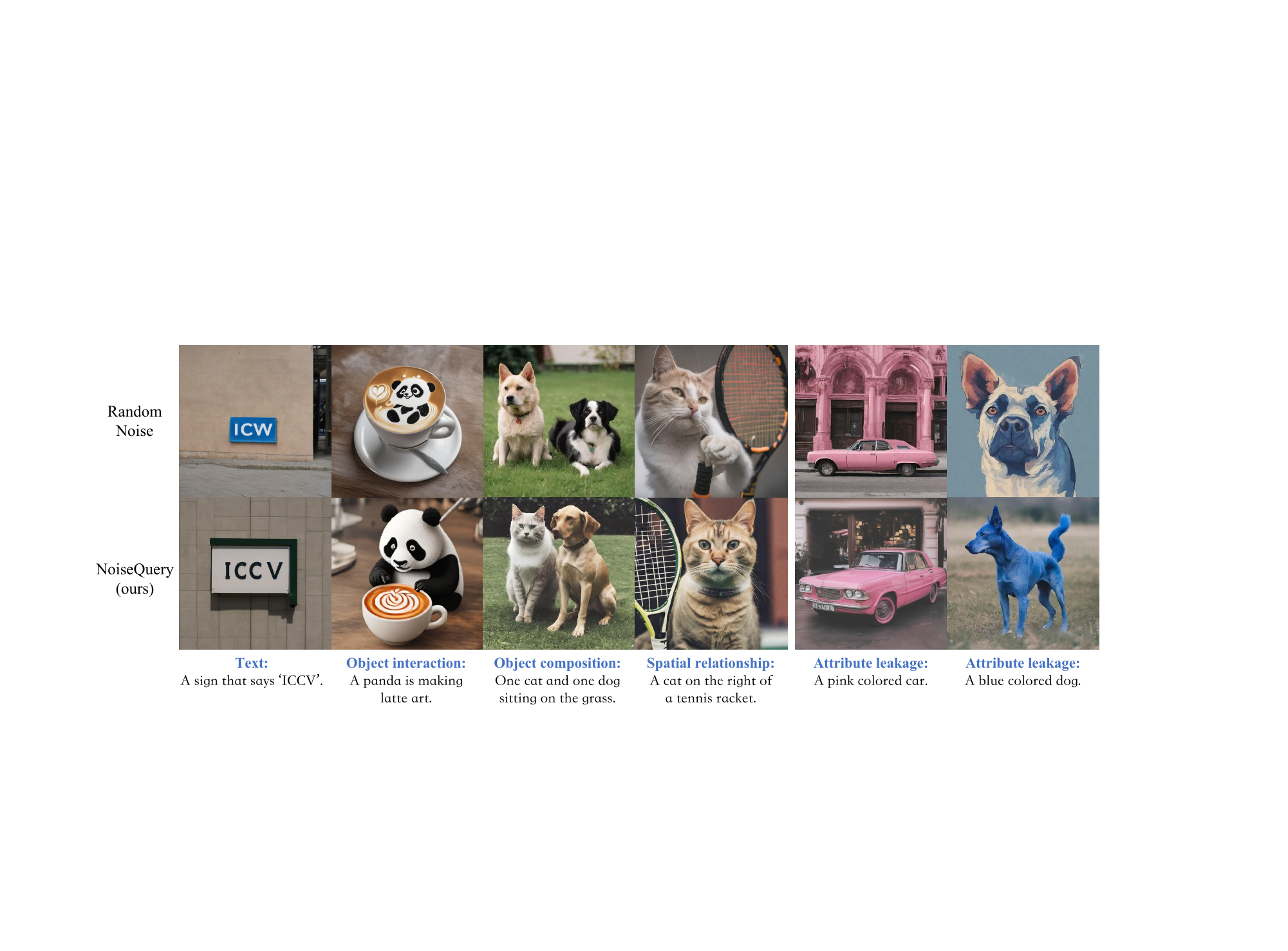}
    \vspace{-6mm}
    \caption{Results on challenging scenarios, showcasing how our method enhances performance in complex image generation tasks. The example shown uses CLIP features for semantic queries.}
    \label{fig:challenging_prompts}
\end{figure*}

\section{Analysis of Generative Posteriors}
\label{sec:sup-sampler}
In this work, we define the generative posterior as the deterministic output of DDIM sampling from a specific initial noise with a NULL text prompt. As analyzed in \cref{sec:initial_noise_analisis}, generative posteriors reveal the hidden features of the initial noise and exhibit model-agnostic behavior.

\begin{figure*}
    \centering
    \includegraphics[width=.85\linewidth]{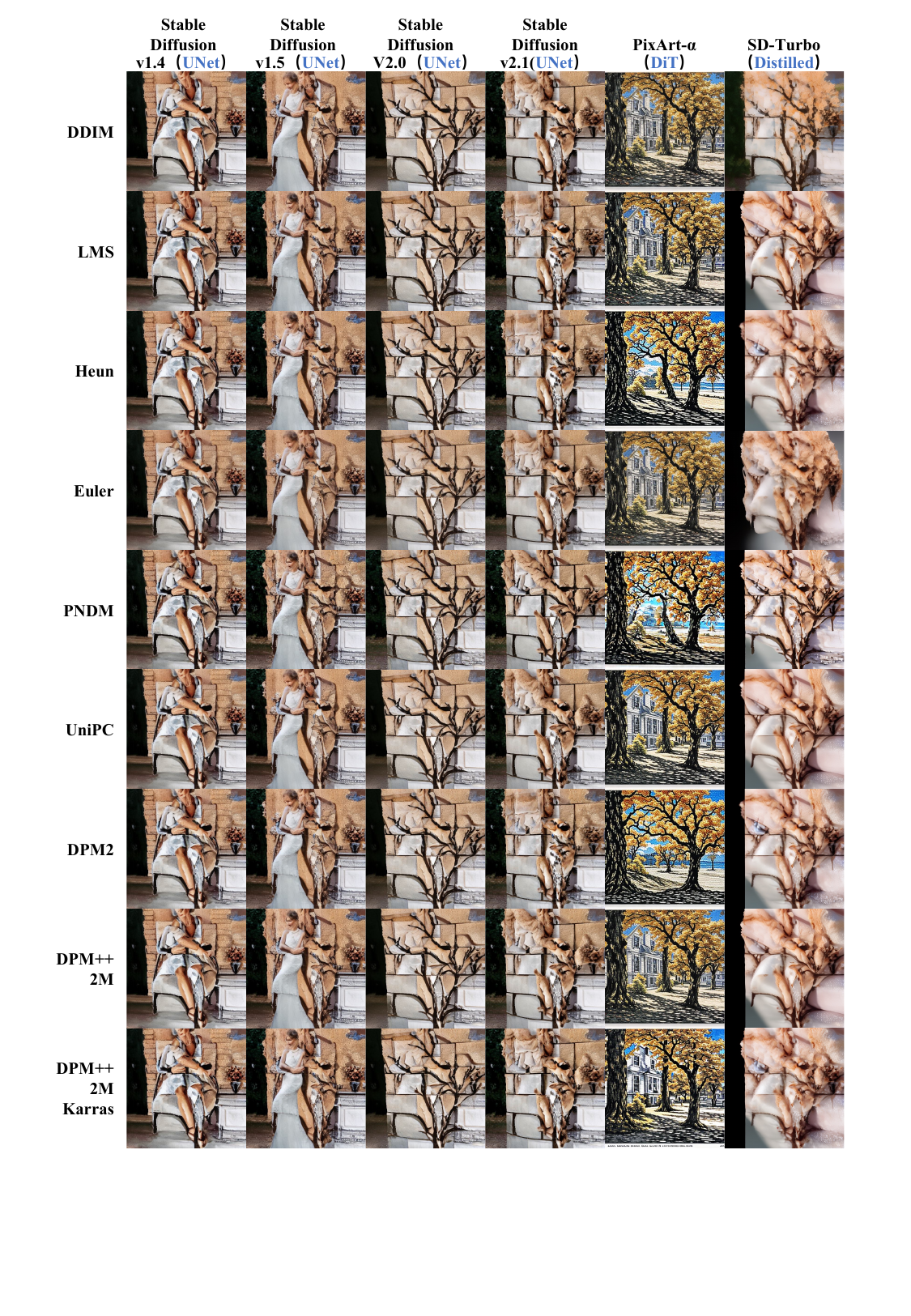}
    \caption{Generative posteriors obtained from the same initial noise using different models (columns) and samplers (rows).}
    \label{fig:sup-sampler}
\end{figure*}

To further investigate the role of samplers, we compare generative posteriors from the same initial noise across multiple models (Stable Diffusion~\cite{rombach2022high-ldm} v1.4, v1.5, v2.0, v2.1, SD-turbo~\cite{sauer2025adversarial-sdturbo} and PixArt-\(\alpha\)~\cite{chen2023pixartalpha}) and samplers (DDIM~\cite{song2020denoising-ddim}, LMS, Heun, Euler, PNDM~\cite{liu2022pseudo-pndm}, UniPC~\cite{zhao2024unipc}, DPM2~\cite{karras2022elucidating}, DPM++ 2M~\cite{dpm++}, and DPM++ 2M Karras\cite{dpm++,karras2022elucidating}). As illustrated in \cref{fig:sup-sampler}, the generated images display remarkable consistency across different models and sampling strategies.
This universal consistency enables the creation of a model-agnostic noise library, where each noise sample is associated with specific latent features and can be seamlessly reused across models and samplers. These findings underline the versatility of leveraging initial noise in a wide range of generative tasks.

\section{Extended Analysis of High-Level Semantics}
\label{sec:sup-more_semantics}

\subsection{Zero-Shot Transferability}\label{sec:sup-zero_shot_transfer}
Our findings show that implicit information in initial noise remains consistent across different models, regardless of their architectures, as discussed in \cref{sec:initial_noise_analisis}.
To validate this, we directly apply our noise library built on Stable Diffusion v2.1 to various different models for generation in a zero-shot manner.
As shown in \cref{Tab:model_transfer}, our approach consistently improves performance across all models, demonstrating its robustness and broad applicability.
This highlights the generalizability of our method, enabling seamless integration with any diffusion model. 

Additionally, NoiseQuery provides a goal-aligned initial noise while users can freely choose various schedulers for generation.
As shown in~\cref{Tab: transfer-scheduler}, a library built with DDIM generalizes well to other schedulers.

\begin{table}[t]
\footnotesize
\centering
\begin{tabular}{lccc}
\toprule
\textbf{Model} & \textbf{Noise} & \textbf{PickScore} & \textbf{CLIPScore}  \\
\midrule
\multirow{2}{*}{SD v2.1 $\rightarrow$ SD v1.4} & Random  & 21.40 & 31.16 \\
 & NoiseQuery & \textbf{21.47} & \textbf{31.26} \\
\midrule
\multirow{2}{*}{SD v2.1 $\rightarrow$ SD v1.5} &  Random  & 21.41 & 31.08 \\
 & NoiseQuery &  \textbf{21.49} & \textbf{31.29} \\
\midrule
\multirow{2}{*}{SD v2.1 $\rightarrow$ SD v2.0} & Random  & 21.63 & 31.60  \\
 & NoiseQuery & \textbf{21.68} &\textbf{ 31.75} \\
\midrule
\multirow{2}{*}{SD v2.1 $\rightarrow$ PixArt-\(\alpha\)}  & Random  & 22.24 & 31.48 \\
 & NoiseQuery &  \textbf{22.32}& \textbf{31.67} \\
\midrule
\multirow{2}{*}{SD v2.1 $\rightarrow$ SD-Turbo}  & Random  & 22.07 &31.51 \\
 & NoiseQuery &  \textbf{22.19} & \textbf{31.70} \\
\bottomrule
\end{tabular}
\caption{Zero-shot transferability from SD v2.1~\cite{rombach2022high-ldm} to various generative models on MSCOCO~\cite{lin2014microsoft-coco} using BLIP features as semantic query. \emph{NoiseQuery} outperforms random noise consistently, proving the initial noise is a universal implicit assistant.}
\label{Tab:model_transfer}
\end{table}

\begin{table}[t]
\centering
\large
\resizebox{\linewidth}{!}{
\begin{tabular}{llcccc}
\toprule
\textbf{Scheduler} & \textbf{Method} & \textbf{ImageReward${\uparrow}$} & \textbf{PickScore${\uparrow}$} & \textbf{HPS v2${\uparrow}$} & \textbf{CLIPScore${\uparrow}$} \\
\midrule
PDNM & SD v2.1 / \textcolor{noisequerycolor}{+NoiseQuery} & 0.10 / \textbf{0.26} & 21.31 / \textbf{21.42} & 24.71 / \textbf{25.23} & 30.91 / \textbf{31.67} \\
DDIM & SD v2.1 / \textcolor{noisequerycolor}{+NoiseQuery} & 0.12 / \textbf{0.26} & 21.30 / \textbf{21.45} & 24.72 / \textbf{25.17} & 31.18 / \textbf{31.71} \\
LMS & SD v2.1 / \textcolor{noisequerycolor}{+NoiseQuery} & 0.07 / \textbf{0.27} & 21.31 / \textbf{21.48} & 24.66 / \textbf{25.37} & 30.83 / \textbf{31.76} \\
Heun & SD v2.1 / \textcolor{noisequerycolor}{+NoiseQuery} & 0.09 / \textbf{0.27} & 21.34 / \textbf{21.48} & 24.66 / \textbf{25.29} & 30.93 / \textbf{31.23} \\
Euler & SD v2.1 / \textcolor{noisequerycolor}{+NoiseQuery} & 0.12 / \textbf{0.22} & 21.35 / \textbf{21.38} & 24.61 / \textbf{25.12} & 31.23 / \textbf{31.54} \\
DPM2 & SD v2.1 / \textcolor{noisequerycolor}{+NoiseQuery} & 0.09 / \textbf{0.27} & 21.31 / \textbf{21.42} & 24.66 / \textbf{25.37} & 30.81 / \textbf{31.79} \\
DPM++ 2M Karras & SD v2.1 / \textcolor{noisequerycolor}{+NoiseQuery} & 0.10 / \textbf{0.27} & 21.36 / \textbf{21.44} & 24.77 / \textbf{25.32} & 30.83 / \textbf{31.79} \\
DPM++ 2M SDE Karras & SD v2.1 / \textcolor{noisequerycolor}{+NoiseQuery} & 0.23 / \textbf{0.28} & 21.42 / \textbf{21.59} & 23.32 / \textbf{25.68} & 31.20 / \textbf{31.23} \\
\bottomrule
\end{tabular}}
\vspace{-2mm}
\caption{Zero-shot transferability from DDIM to various samplers on DrawBench~\cite{saharia2022photorealistic-drawbench}.}
\label{Tab: transfer-scheduler}
\end{table}

\subsection{Complex Image Generation.}
As shown in \cref{fig:challenging_prompts}, our method improves the model's performance in challenging scenarios like visual text generation, object interaction, object composition, and spatial relationship, while reducing attribute leakage.
For example, when generating a ``blue dog", a random noise might cause the model to adaptively infuse the blue into the entire scene. This is because the model tries to enforce semantic consistency globally, which can destabilize the generation process, leading to unintended attribute spillover.
In contrast, our approach selects noise that already contains target features (e.g., objects, layouts), minimizing the need for global adjustments. This preserves object attributes, enhances semantic consistency, and prevents attribute leakage.

\subsection{Different CFG Scale Analysis}\label{sec:sup-scales}

We provide a comprehensive analysis of our method’s performance across various classifier-free guidance (CFG) scales on 10k MSCOCO prompts in \cref{fig:coco_curve}. 
Remarkably, even at low CFG scales, our approach achieves comparable or superior results compared to the baseline at much higher scales. 
This demonstrates that the selected noise samples effectively align with text prompts, reducing generation difficulty and eliminating the need for excessively high guidance scales that often cause over-saturation and instability.

\begin{figure*}
    \centering
    \includegraphics[width=.9\linewidth]{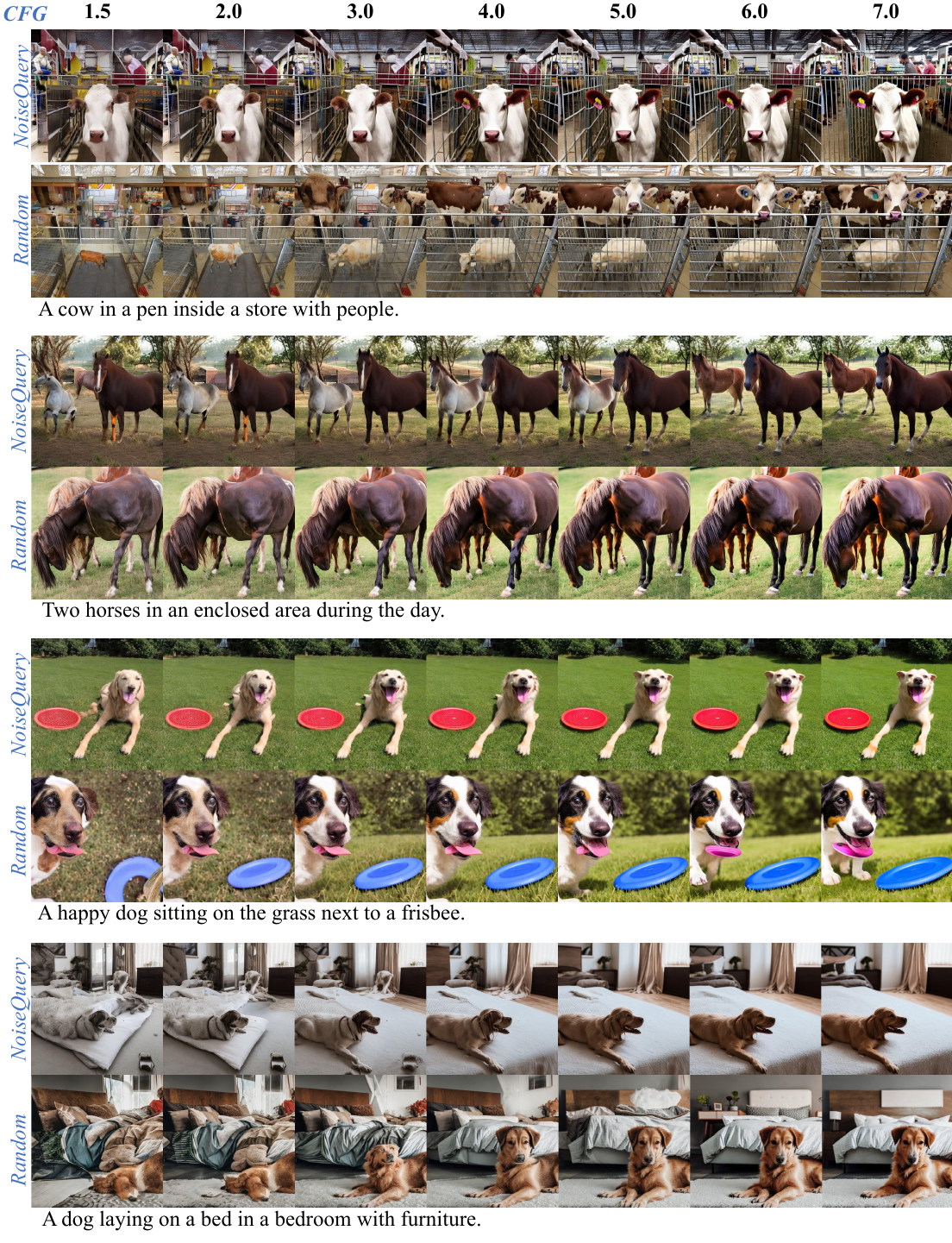}
    \caption{Visual comparison of generated images at different CFG scales (1.5, 2.0, ..., 7.0). Our method produces semantically accurate and visually coherent images at low scales, while random noise often fails to generate meaningful or semantically consistent outputs. This highlights our method’s ability to reduce generation difficulty and improve stability.}
    \label{fig:sup-scales}
\end{figure*}

To complement the quantitative analysis, \cref{fig:sup-scales} visualizes the generated images across different CFG scales. Our method consistently produces semantically accurate and visually coherent outputs even at low scales, while random noise frequently results in failed generations. Even at higher CFG scales, the baseline struggles to maintain stable semantic alignment, whereas our approach achieves robust and reliable performance across most scales.

\section{Implementation Details of Low-Level Visual Properties}\label{sec:sup-lowlevel}
\subsection{Texture}
Texture refers to the small-scale spatial patterns of intensity variation in an image, which can be used to characterize the surface properties of objects. To control texture in generated images, we utilize the Gray-Level Co-occurrence Matrix (GLCM)~\cite{haralick1973textural-glcm}, a commonly used technique for quantifying texture in terms of statistical measures. The GLCM features capture the spatial arrangement and frequency of pixel intensity changes, providing a texture profile for each image. 

For texture-based noise querying, we compare the GLCM features of the reference image with those stored in the noise library. The noise with the closest texture profile (using a distance metric like Euclidean distance) is selected to generate the output image.

\subsection{Shape}
Shape refers to the geometric form or structure of objects in an image, independent of texture or color. To control shape in generated images, we use Hu Moments~\cite{hu1962visual-hu}, which are invariant shape descriptors that capture the overall geometry of an object, regardless of its size, position, or orientation. These moments are derived from the image's spatial distribution of intensity and provide a compact, rotation-invariant representation of the object's shape. 

For shape-based noise querying, the similarity between the reference and each noise sample is measured using Euclidean distance between their Hu Moments vectors. The noise sample with the closest similarity to the reference shape is selected to ensure the generated image maintains the desired shape.

\begin{algorithm}[t]
\caption{Noise Offset for Controlled Color Shifting}
\label{algo:sup-noiseoffset}
\KwIn{
    Diffusion model \( \mathcal{M} \); 
    Initial noise \( \epsilon \); 
    Adjustment parameters \( \delta \) (e.g., brightness, saturation); 
    Color adjustment function \( S(\cdot, \delta) \); 
    DDIM inversion process \( \text{DDIM-Inverse}(\cdot, T) \); 
}
\KwOut{Modified noise \( \epsilon^* \)}

\BlankLine 
\Begin{
    \( \mathcal{I}_{\text{uncond}} \gets \mathcal{M}(\epsilon, T, c = \emptyset) \) \;
    
    \( \mathcal{I}_{\text{adjusted}} \gets S(\mathcal{I}_{\text{uncond}}, \delta) \) \;
    
    \( \epsilon^* \gets \text{DDIM-Inverse}(\mathcal{I}_{\text{adjusted}}, T) \) \;
    
    \Return \( \epsilon^* \) \;
}
\end{algorithm}

\begin{figure}[t]
    \centering
    \includegraphics[width=\linewidth]{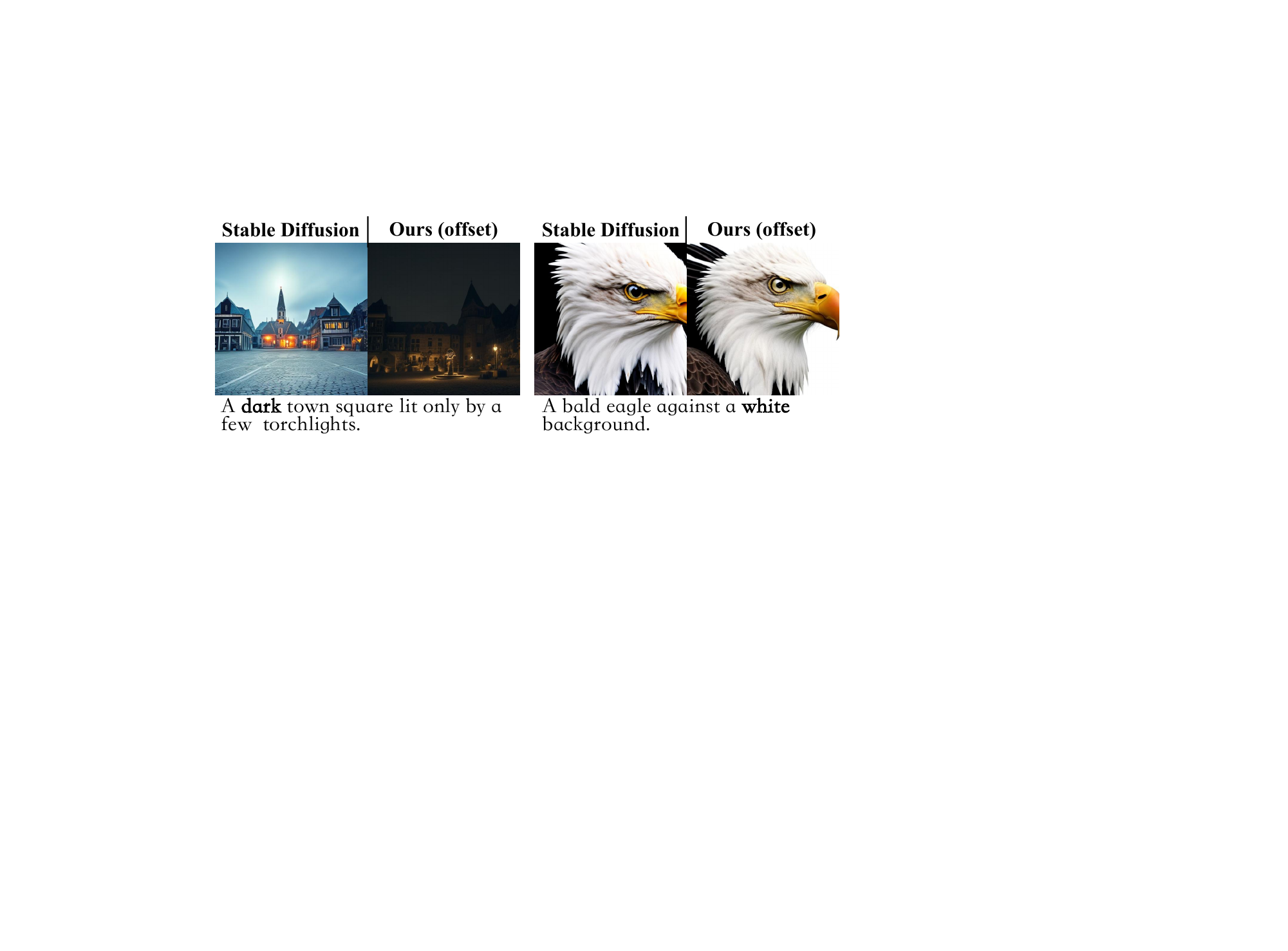}
    \vspace{-6mm}
    \caption{Unlike the original Stable Diffusion, which produces images with medium brightness, our \emph{NoiseQuery} (offset) expands the range to include both very bright and very dark samples.}
    \label{fig:brightness}
\end{figure}
\subsection{Sharpness}

Sharpness is a key low-level visual property that relates to the level of detail and clarity in an image, particularly the prominence of high-frequency components, such as edges and fine textures.  To control sharpness in the generated images, we focus on measuring High-Frequency Energy (HFE). HFE quantifies the amount of high-frequency content (i.e., fine details and sharp edges) in an image.  Higher HFE corresponds to sharper, more detailed images, while lower HFE indicates softer, blurrier images. The noise sample with the higher HFE (based on sorting) is chosen, ensuring the generated image has the desired sharpness.

\subsection{Noise Offset for Controlled Color Shifting}\label{sec:sup-noiseoffset}
Beyond directly selecting noise, we also propose a method to introduce subtle color shifts by adjusting generative posteriors, preserving semantic content while allowing controlled color variations.
We show the pseudo-code in ~\cref{algo:sup-noiseoffset}
Specifically, we first generate an unconditional image from the initial noise and then apply subtle color adjustments, such as changes in brightness or saturation. After the color changes are made, we reverse the adjusted image back into the noise space via DDIM inversion. This results in a modified noise that consistently carries the desired color shifts, ensuring uniform color variations across different prompts while maintaining the underlying semantic content.

This helps address the limitation of SD~\cite{rombach2022high-ldm} in generating very bright or dark images, caused by noise distribution discrepancies during training and inference~\cite{lin2024common-noiseschedule}. As shown in \cref{fig:brightness}, subtly adjusting the brightness of generative posteriors can shift the inherent tendency of the initial noise, enabling the generation of images with a wider range of brightness during inference.
shifts the noise distribution, enabling a wider range of brightness in generated images.

\end{document}